\documentclass[journal]{IEEEtran}
\usepackage[utf8]{inputenc}
\usepackage{tabu}
\usepackage{times}
\usepackage[T1]{fontenc}
\usepackage{graphicx}
\graphicspath{{images/}}
\usepackage{algpseudocode}
\usepackage{adjustbox}
\usepackage{hyperref}

\usepackage{subcaption}
\usepackage{algorithm}
\usepackage{stfloats}
\usepackage{float}
\fnbelowfloat

\usepackage{url}
\usepackage{amsmath,amssymb,amsfonts}
\usepackage{textcomp}
\usepackage{multirow}
\usepackage{booktabs}
\ifCLASSINFOpdf
\else
\fi
\begin{document}
%
\title{Technical Report on Subspace Pyramid Fusion Network for Semantic Segmentation}

	

%

\author{Mohammed A. M. Elhassan\textsuperscript{1}, Chenhui Yang\textsuperscript{2}, Chenxi Huang\textsuperscript{2}
        and Tewodros Legesse Munea\textsuperscript{2}\\
        \hspace{18cm}
         \textsuperscript{1}School of Computer Science, Zhejiang Normal University,688 Yingbin Road,Jinhua,Zhejiang,China \\
        \textsuperscript{2}School of Informatics, Xiamen University,Xiamen, China,361005
\thanks{Email address: \url{mohammedac29@zjnu.edu.cn}(Mohammed A. M. Elhassan).}
}
\maketitle

\begin{abstract}
The following is a technical report to test the validity of the proposed Subspace Pyramid Fusion Module (SPFM) to capture multi-scale feature representations, which is more useful for semantic segmentation. In this investigation, we have proposed the Efficient Shuffle Attention Module(ESAM) to reconstruct the skip-connections paths by fusing multi-level global context features. Experimental results on two well-known semantic segmentation datasets, including Camvid and Cityscapes, show the effectiveness of our proposed method. The source code is available at \footnote{\url{https://github.com/mohamedac29/SPFNet}}
\end{abstract}


%
\IEEEpeerreviewmaketitle


\section{Introduction}
\label{secI:introduction}
\IEEEPARstart{S}{Semantic} Semantic segmentation is an essential high-level topic in computer vision and has been widely used in various challenging problems such as, medical diagnosis, \cite{ronneberger2015u,saha2018her2net}, autonomous vehicles  \cite{siam2018comparative}, and scene analysis \cite{zhao2018psanet,zhao2017pyramid}. Unlike image classification, which aims to classify the whole image, semantic segmentation predicts the per-pixel class for each image content. The current frontiers in semantic segmentation methods are driven by the success of deep convolution neural networks, specifically the fully convolution network (FCN) Framework \cite{long2015fully}. In the original FCN approach to obtain larger receptive fields, they increase the depth of the network with more convolutional and downsampling layers. However, simply increasing the number of downsampling operations leads to reduced feature resolution, which causes spatial information loss. On the other hand, the increase in the number of convolutional layers adds more challenges to network optimization. 
\begin{figure}
    \centering
    \includegraphics[width=0.9\linewidth]{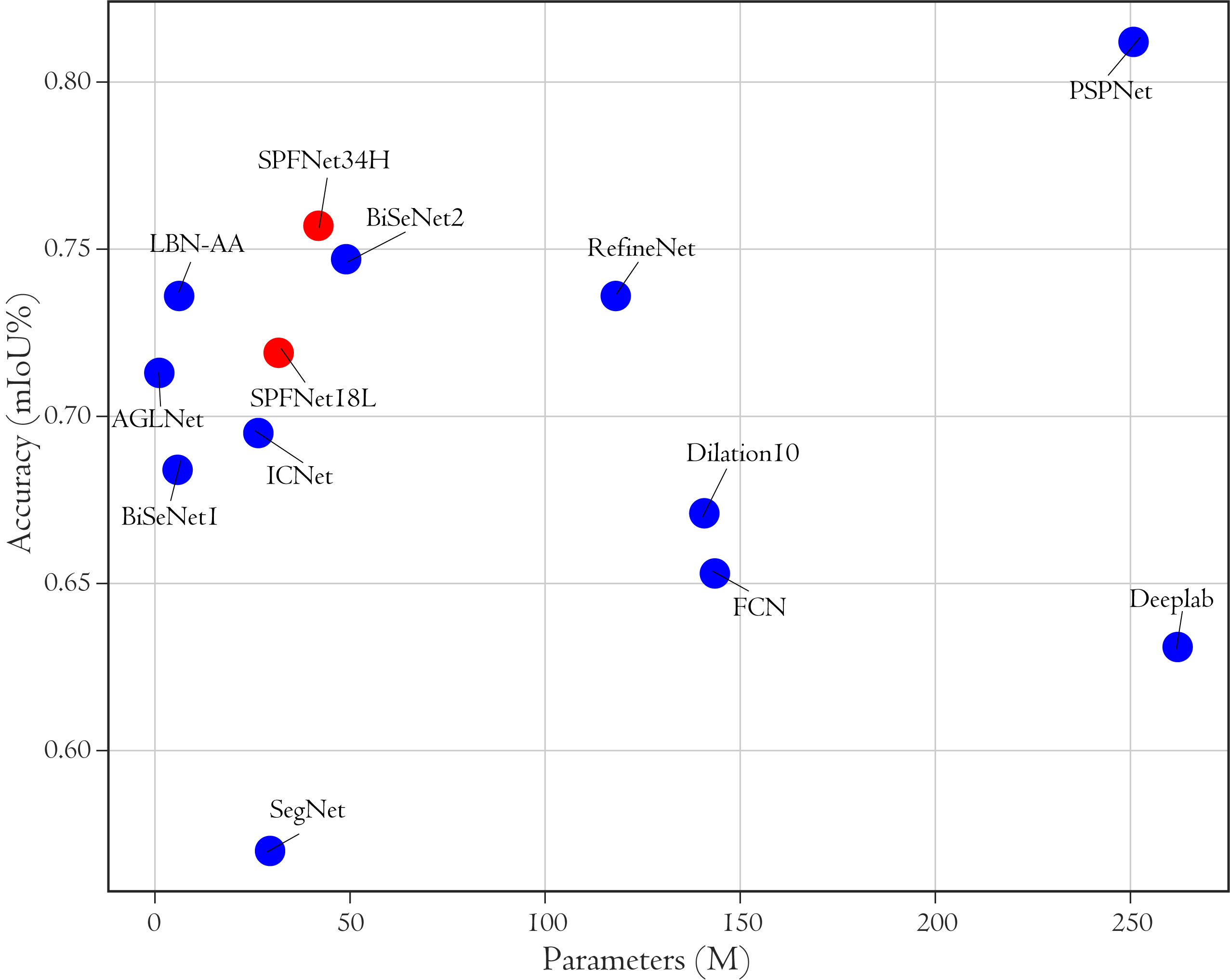}
   \caption{Accuracy-Speed performance comparison on the Cityscapes test set. Our methods are presented in red dots while other methods are presented in blue dots. Our approaches achieve state-of-the-art speed-accuracy trade-off}
    \label{fig:accuracy_speed}
\end{figure}

Over the years, various approaches have been proposed to tackle the drawbacks of the typical FCN-based architecture. Here we listed three categories of methodologies as follows: (i) increasing the resolution of feature maps or maintaining a high-resolution feature across stages, e.g., through decoder network \cite{ronneberger2015u,badrinarayanan2017segnet,zhou2018unet++}, dilated convolutions  \cite{yu2015multi,chen2017deeplab} or high resolution \cite{sun2019deep}. (ii) improve the segmentation performance with the subsequent approaches, e.g., PSPNet \cite{zhao2017pyramid}, DeepLab \cite{chen2017deeplab,chen2018encoder}, PAN \cite{li2018pyramid} DenseASPP \cite{yang2018denseaspp} , RefineNet \cite{lin2017refinenet}, PPANet \cite{elhassan2021ppanet}, and ParseNet \cite{liu2015parsenet}. These previous networks enlarge the receptive field to exploit the rich contextual information. Nevertheless, the latter solution is infeasible for real-time applications because it requires heavy networks and expensive computation. (iii) some networks such as  \cite{elhassan2021dsanet,yu2018bisenet} have added spatial encoding path to preserve the spatial details
To perform fast segmentation with light computation cost, small size and satisfactory segmentation accuracy, there are certain design philosophy that could be followed: (i) incorporate a pre-trained lightweight classification networks\cite{iandola2016squeezenet,howard2017mobilenets,ma2018shufflenet,chollet2017xception} as a backbone to construct segmentation architecture such as BiSeNet \cite{yu2018bisenet}, STDC-Seg \cite{fan2021rethinking}. (ii) design a building block that is suitable for low computation cost \cite{li2019dabnet,romera2017erfnet}. (iii) put into consideration the downsampling strategy and the depth of the network \cite{li2019dfanet}. (iv) combine the low-level and high-level features using the multi-path framework \cite{yu2018bisenet,yu2021bisenet,elhassan2021dsanet}. in addition to that, restricting the input image size plays an important part in increasing the inference speed.
\begin{figure*}
	\centering
	\begin{subfigure}[b]{0.32\textwidth}
		\includegraphics[width=5.8cm,height=6cm]{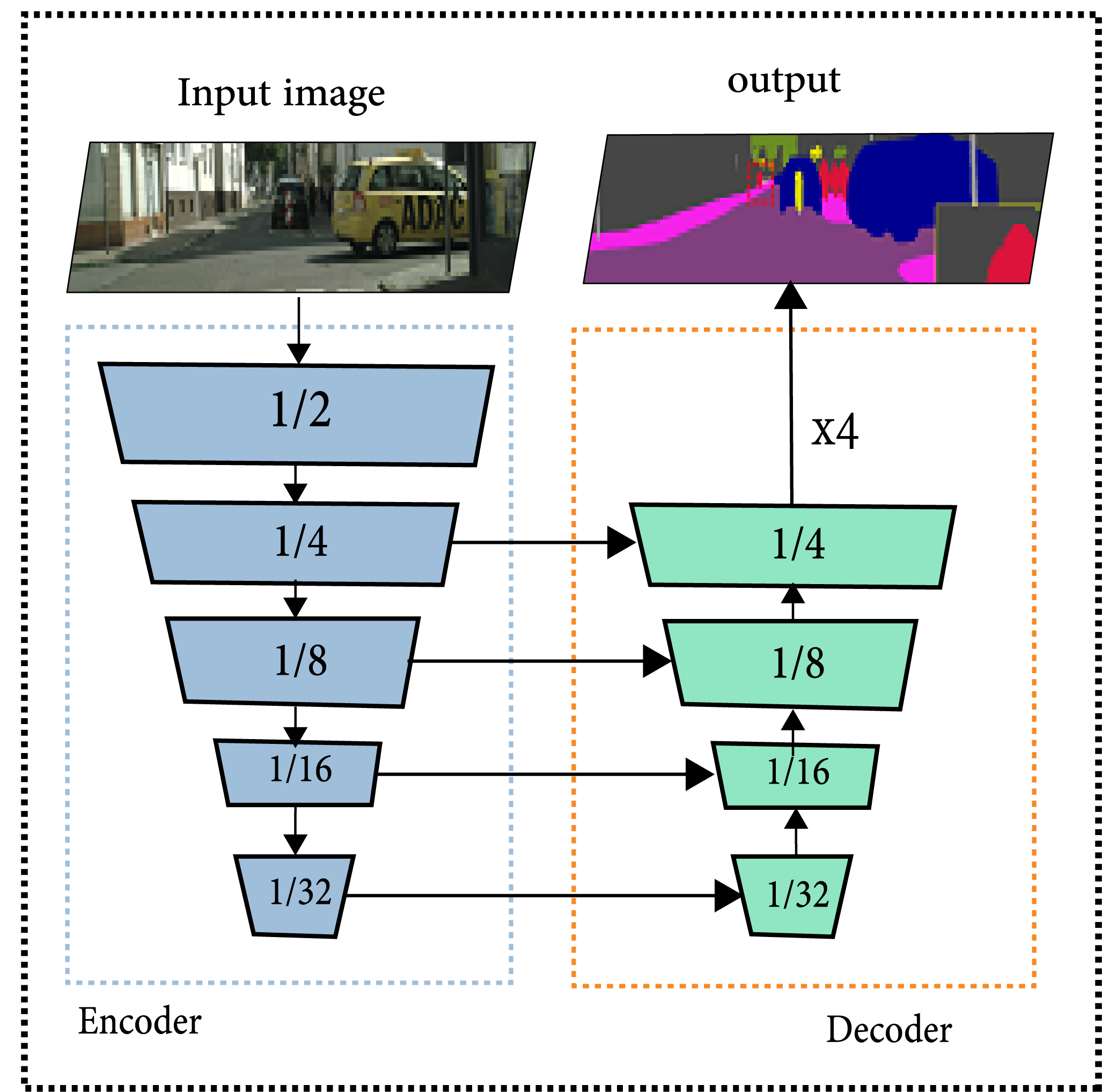}
		\caption{Encoder-Decoder Structure}
		\label{fig1:encoder_decoder}
	\end{subfigure} \hspace{1mm}
	\begin{subfigure}[b]{0.32\textwidth}
		\includegraphics[width=5.8cm,height=6cm]{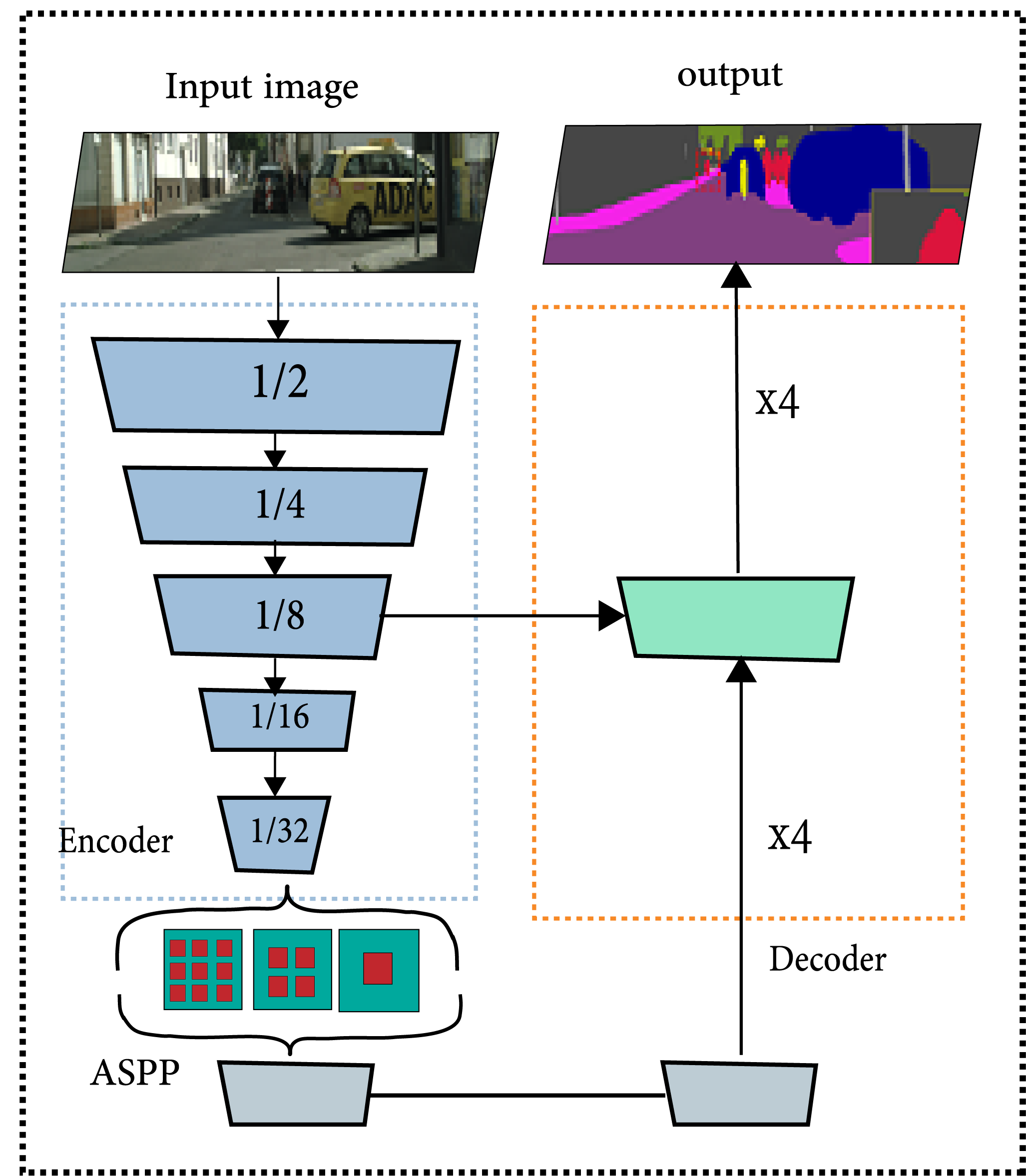}
		\caption{Encoder-Decoder with Atrous Conv}
		\label{fig2:atrous_convolution}
	\end{subfigure}\hspace{1mm}
	\begin{subfigure}[b]{0.32\textwidth}
	    \includegraphics[width=5.8cm,height=6cm]{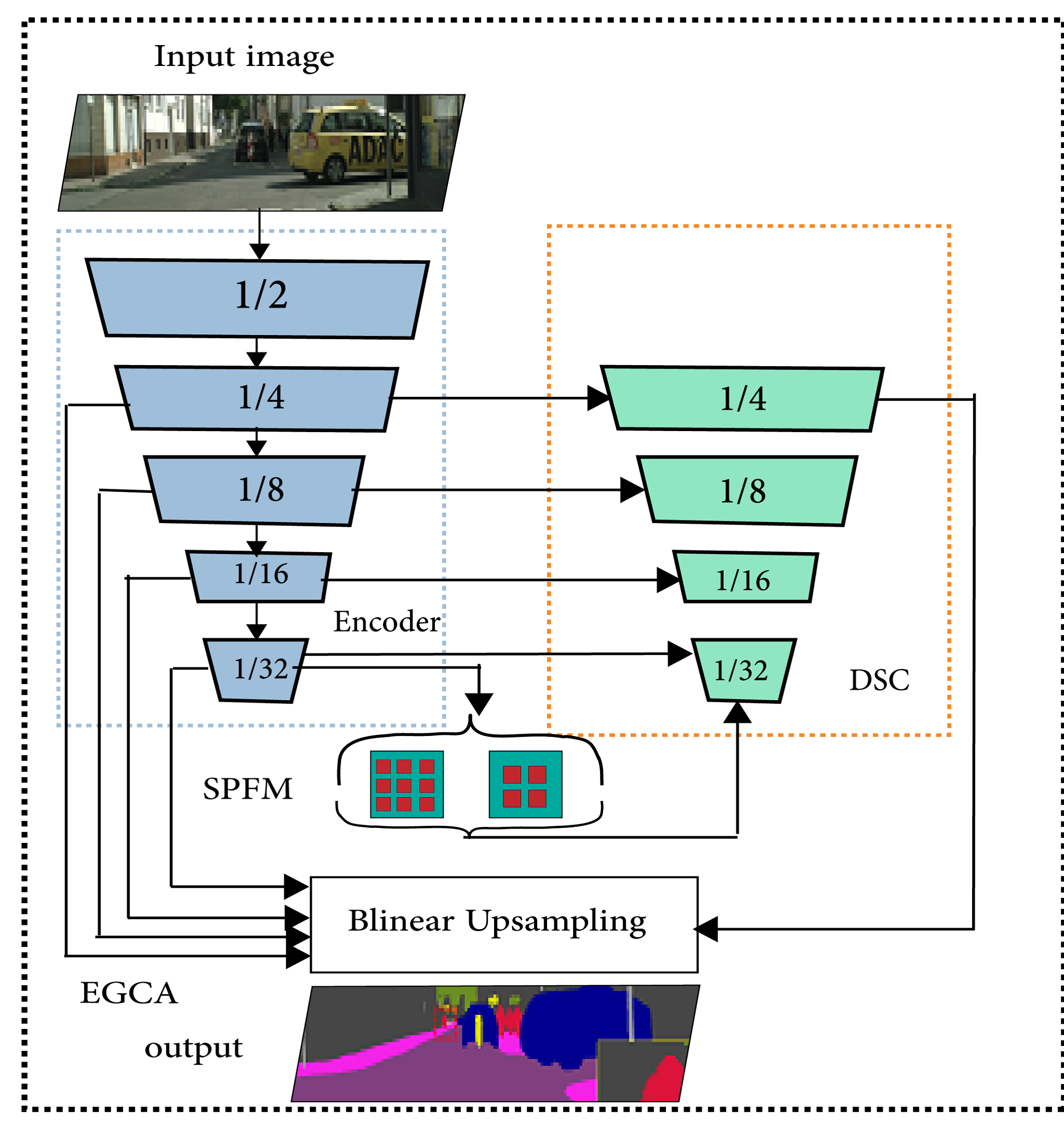}
	    \caption{Encoder-Decoder with SPFM}
	    \label{fig3:fpn}
	\end{subfigure}\hspace{1mm}
	\caption{A comparison of an important semantic segmentation architectures. (a) encoder-decoder structure.(b) encoder-decoder structure with atrous convolutions.(c) our network, encoder-decoder structure with atrous convolutions and context aggregation module.}
	\label{fig:networks_comparison}
\end{figure*}

\begin{figure*}
	\centering
	\includegraphics[width=0.7\linewidth]{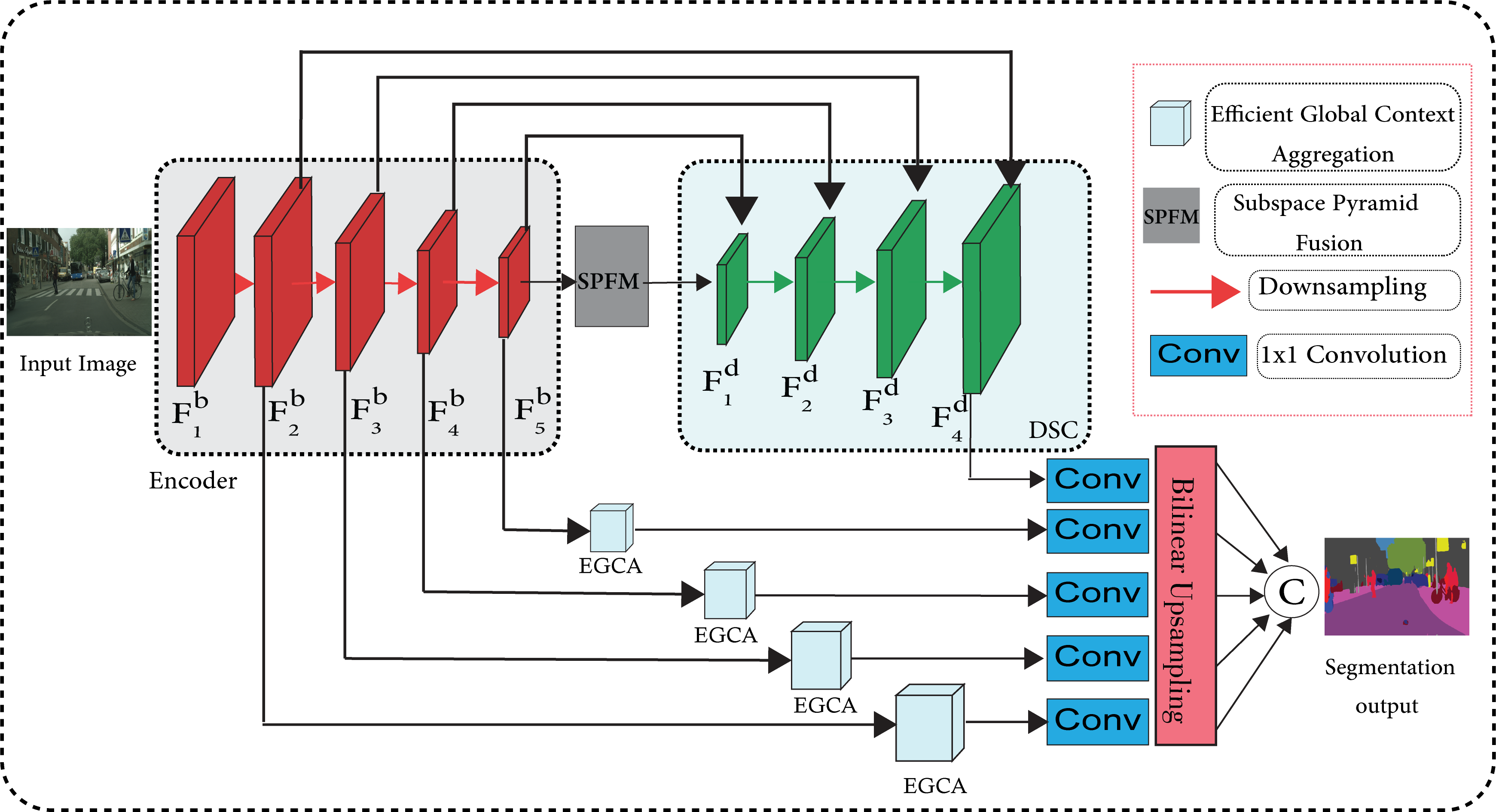}
	\caption{An overview of the proposed Subspace Pyramid Fusion Network (SPFNet).}
	\label{spfnet}
\end{figure*}

Multi-scale information is the most crucial problem that should be considered to improve the performance of semantic segmentation networks. Typically, a class of targets may exist at different scales in the image, and a well-designed model should be able to extract this property, which leads to better semantic segmentation accuracy. In this paper, we propose a Subspace Pyramid Fusion Module (SPFM) to improve the multi-scale feature learning in semantic segmentation. In particular, the SPFM generates global contextual information in each feature subspace based on parallel Reduced Pyramid Pooling submodule (RPP). RPP submodule constructed from three components: dilated convolution or atrous convolution, subpixel convolution, and average pooling. Two dilated convolutions were used for each reduced pyramid pooling level to capture object features with various receptive fields, followed by subpixel convolution instead of the original upsampling method to unify the feature resolution. To address the problems of resolution loss via downsampling in the encoding part, we develop the efficient shuffle attention module (ESAM) with two branches. Through the lower branch, i.e., ESAM integrates the shuffle attention block to facilitate cross-channel information communication for feature maps from different encoder levels and reduces channel redundancy. Finally, a Decoder-based Subpixel Convolution (DSC) is proposed to improve category pixel localization and recover object details. The SPFM module is embedded in the center between the encoder and the decoder to help select the suitable receptive field for objects with different scales by fusing the multi-scale context information. Based on these descriptions, we design the Subspace Pyramid Fusion Network (SPFNet), see Figure\ref{spfnet}. The proposed SPFNet is validated with two challenging semantic segmentation benchmarks, Camvid \cite{brostow2009semantic} and Cityscapes \cite{cordts2016cityscapes}. Figure. \ref{fig:accuracy_speed} shows the accuracy and the speed comparison of different methods on Cityscapes dataset.

Furthermore, many existing frameworks \cite{ghiasi2019fpn,liu2018path,tan2020efficientdet,kirillov2019panoptic,lin2017feature} have been developed based on FPN \cite{lin2017feature} to exploit the inherent multi-scale feature representation of deep convolutional networks. More specifically, FPN-based architectures combine large-receptive field features, low-resolution- with small-receptive-field features, high-resolution to capture objects at different scales. For instance, \cite{kirillov2019panoptic,lin2017feature} utilize the lateral path to fuse adjacent features in a top-down manner. This way FPN strengthen the learning of multi-scale representation.Until now, feature pyramid network based approaches have achieved state-of-arts performance in object detection. 
Our main contributions are summarized as follows:\\
\begin{enumerate}
	\item A novel Subspace Pyramid Fusion (SPFM) module is proposed to learn multi-scale representations.
	\item We introduce the Efficient Shuffle Attention Module(ESAM) module, which utilizes channel shuffle operation to integrate the complementary global context 
	\item  We proposed an architecture based on SPFM and ESAM modules that can be easily applied for road scene understanding. Furthermore, these modules can be plugged in any network
\end{enumerate}

\section{Related Work}
\subsection{Efficient Network Designs}
In recent literature, real-time model design has become an essential part of computer vision research development. Several lightweight architectures were introduced as a trade-off between accuracy and latency SqueezeNet \cite{iandola2016squeezenet}, MobileNet \cite{howard2017mobilenets,sandler2018mobilenetv2}, ShuffleNet \cite{zhang2018shufflenet,ma2018shufflenet},and Xception \cite{chollet2017xception}. SqueezeNet reduces the number of parameters through the squeeze and expansion module. MobileNetV1 implemented depth-wise separable convolutions to reduce the number of parameters. MobileNetV2 proposed an inverted residual block to alleviate the effect of depth-wise separable convolutions. ShuffleNet and Xception utilize group convolution to reduce FLOPs. ResNet \cite{he2016deep} adopted residual blocks to obtain outstanding performance. These models were designed primarily for image classification tasks. Later on, incorporated to work as backbones for semantic segmentation networks to achieve real-time semantic segmentation.
\subsection{Real-time Segmentation}
Real-time semantic segmentation methods aim to predict dense pixels accurately while maintaining high inference speed. Over the years there are different approaches have been proposed in this regard. ENet \cite{paszke2016enet}, ERFNet \cite{romera2017erfnet}, and ESPNet \cite{mehta2018espnet,mehta2019espnetv2} proposed compact semantic segmentation networks that utilize lightweight backbones. LEDNet \cite{wang2019lednet}, DABNet \cite{li2019dabnet} introduced a lightweight model to address the computation burden that limits the usage of deep convolutional neural networks in mobile devices. The former used asymmetric encoder-decoder-based architecture and implemented channel split and shuffle operations to reduce the computation cost. The latter structured a depth-wise symmetric building block to jointly extract local and global context information.  FDDWNet \cite{liu2020fddwnet} utilized factorized depth-wise separable convolutions to reduce the number of parameters and the dilation to learn multi-scale feature representations.
ICNet \cite{zhao2018icnet} proposes a compressed version of  PSPNet based on an image cascade network to increase the semantic segmentation speed. DFANet \cite{li2019dfanet} obtains the multi-scale feature propagation by utilizing a sub-stage and sub-network aggregation.Tow-pathway networks such as  BiSeNetV2 \cite{yu2021bisenet} and DSANet \cite{elhassan2021dsanet} Fast-SCNN \cite{poudel2019fast} introduce a shallow detailed branch to extract the spatial details and a deep semantic branch to extract the semantic context. This paper propose a lightweight model that utilizes (ResNet18 or ResNet34) as backbones. 

\section{Methodology}
In this section, we first present an overview of the SPFNet, and then elaborate the mechanism of different modules that used to construct this network.

\subsection{Overview}
Figure \ref{spfnet} illustrates the architectural framework of SPFNet, designed within a conventional encoder-decoder paradigm. The encoder is responsible for downsampling the input data to acquire diverse multi-scale features, while the decoder focuses on reconstructing high-level features from low-level ones, facilitating pixel-level semantic predictions. SPFNet initiates the process by taking an input image and encoding its features through a widely-used multi-scale backbone, specifically ResNet34 \cite{he2016deep} ,with hierarchical representations $[F_{1}^{b},F_{2}^{b},F_{3}^{b},F_{4}^{b},F_{5}^{b}]$. The notation $[F_{2}^{egca},F_{3}^{egca},F_{4}^{egca},F_{5}^{egca}]$ denotes the rich context information extracted from the feature hierarchy within the Efficient Shuffle Attention Module (ESAM) (see section \ref{sec:shuffle_attention}).
\begin{figure*}[t]
	\centering
	\begin{subfigure}[b]{0.48\textwidth}
		\includegraphics[width=8.2cm,height=5.4cm]{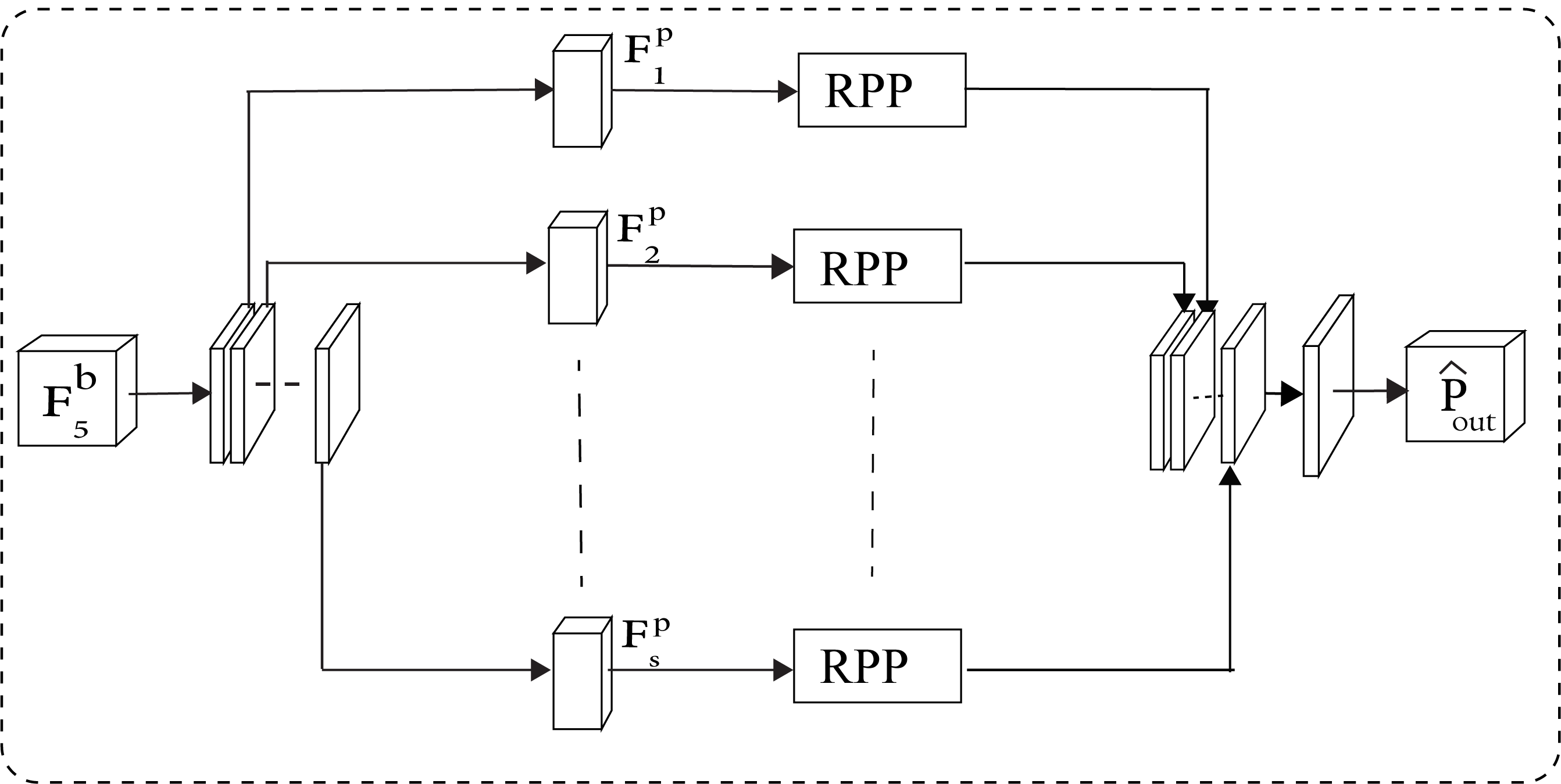}
		\caption{Subspace pyramid fusion module (SPFM).}
		\label{subfig:spfm}
	\end{subfigure} \hspace{1.5mm}
	\begin{subfigure}[b]{0.48\textwidth}
		\includegraphics[width=8cm,height=5.4cm]{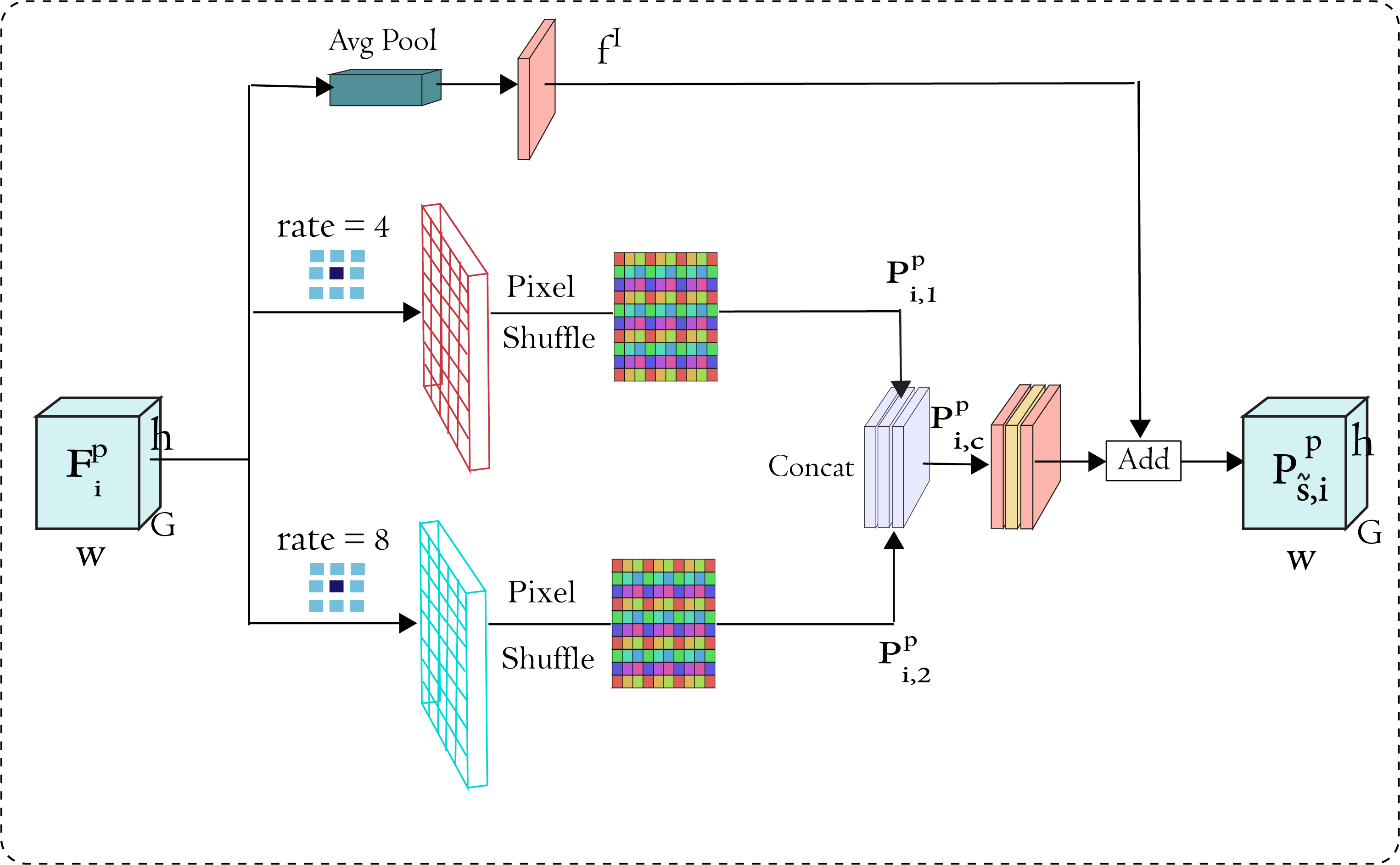}
		\caption{Reduced pyramid pooling module (RPP).}
		\label{subfig:rpp}
	\end{subfigure}\hspace{1.5mm}
	\caption{Illustration of the (a) Reduced pyramid pooling module (RPP). (b) Subspace Pyramid Fusion (SPFM) module. SPFM extracts the multi-scale information in each sub-features using RPP module, then transmit the information to decoder.}
	\label{fig:sub_pyramid}
\end{figure*}

\subsection{Subspace Pyramid Fusion Module (SPFM)}
\label{sec:subspace_pyramid}
In the realm of semantic segmentation research \cite{chen2017deeplab, gu2019net, li2018pyramid}, it has been substantiated that augmenting receptive fields is advantageous for the semantic segmentation task. A conventional encoder-decoder architecture involves the extraction of global context information by the encoder from the input image, encompassing the adjacent and class characteristics of the object. However, the transmission of information to shallower layers is known to attenuate the extraction of context information owing to downsampling processes. Motivated by this observation and drawing inspiration from prior works such as \cite{chen2017deeplab} and \cite{saini2020ulsam}, we introduce a pioneering Subspace Pyramid Fusion Module (SPFM) in this subsection.

Within the framework of SPFM, each subspace feature undergoes a singular reduced pyramid pooling (RPP) treatment. In contrast to methodologies presented in \cite{chen2017deeplab} and \cite{xie2018vortex}, SPFM, illustrated in Figure \ref{fig:sub_pyramid}, diverges in its approach to collecting multi-scale contextual information. Rather than directly fusing dilated convolutions, SPFM initiates the generation of multi-scale features for each split using a reduced pyramid pooling module. The RPP is constructed with only two parallel dilated convolutions, employing dilation rates of (4,8). Subsequently, pixel shuffle is employed to enhance feature resolution. Additionally, adaptive average pooling is integrated with the output from two parallel convolutions to capture complex contextual information. The resulting RPP blocks are concatenated to form the SPFM. In comparison with SPP and ASPP, our module demonstrates superior information acquisition through subpixel convolution and stacked RPP blocks.

Diverging from pyramid-based multi-scale learning approaches such as SPP and ASPP, our SPFM module acquires heightened multi-scale information by learning multiple Reduced Pyramid Pooling blocks (RPP) for each feature map. These RPP blocks are then amalgamated within the Subspace Pyramid Fusion Module. A distinctive characteristic of SPFM, as opposed to \cite{saini2020ulsam}, lies in the approach to information aggregation. While \cite{saini2020ulsam} integrates attention across different stages in pre-existing compact backbones to facilitate global information learning, SPFM achieves an increase in receptive fields through dilated convolutions and adaptive average pooling to capture global context.

Given the feature map $F_{5}^{b}\in \mathbb{R}^{H\times W\times C}$ The objective of our study is to efficiently capture multi-scale features from the feature map 
$F_{5}^{b}\in \mathbb{R}^{H\times W\times C}$ obtained from the backbone, where 
H and W represent the spatial dimensions of the feature maps, and C denotes the number of channels. In Figure \ref{fig:sub_pyramid}.\subref{subfig:spfm}, we introduce the proposed Subspace Pyramid Fusion Module (SPFM) and its submodule Reduced Pyramid Pooling (RPP) in Figure \ref{fig:sub_pyramid}.\subref{subfig:rpp}. SPFM segments the input feature map 
$F_{5}^{b}$into s splits along the channel dimension.
Here are the details of SPFM and RPP modules.\\
\textbf{Reduced Pyramid Pooling (RPP)}: Each RPP block takes one split from SPFM. First, the input features $F_{i}^{p}$ enter into two identical pathways, each path consists of dilated convolution with kernel size of 3 followed by batch normalization(BN) and Parametric Rectified Linear Unit (PReLU) to generate multi-scale context information from these feature maps. Subsequently, we apply pixel shuffle to obtain $P_{i,1}^{p}$ as in Eq. \ref{p1} and $P_{i,2}^{p}$ Eq. \ref{p2}. Then, we concatenate the output features $P_{i,c}^{p}$ Eq. \ref{spconc1} and refined it with three consecutive convolutions with kernel sizes ($1\times1$), ($3\times3$), and ($1\times1$) respectively, to reduce aliasing effect. Finally, the output is added with feature map of adaptive average pooling and convolution Eq. \ref{avgp} to produce the final reduced pyramid pooling $P_{\tilde{s},i}^{p}$ as shown in Eq \ref{rpp}.

\begin{equation}\label{p1}
P_{i,1}^{p} = \mathbb{PS}(\mathbb{D}_{conv}@4(F_{i}^{p}))
\end{equation}

\begin{equation}\label{p2}
P_{i,2}^{p} = \mathbb{PS}(\mathbb{D}_{conv}@8(F_{i}^{p}))
\end{equation} 

\begin{equation}\label{spconc1}
P_{i,c}^{p} = Concat([P_{i,}^{p},P_{i,2}^{p}])
\end{equation} 

\begin{equation}\label{avgp}
A_{i}^{p}  = f^{1}(APool(F_{i}^{p}))
\end{equation}

where Concat([.]) is the element-wise concatenation.
The reduced pyramid fusion for each split $P_{\tilde{s},i}^{p}$ is computed by the following equation.
\begin{equation}\label{rpp}
P_{\tilde{s},i}^{p} = A_{i}^{p} \oplus P_{i,c}^{p} 
\end{equation} 

In Eq. \ref{p1}, and Eq. \ref{p2}, $\mathbb{D}_{conv}@$ refers to dilated convolution layer with kernel size of $3\times3$. In Eq\ref{avgp} APool(.), is the adaptive average pooling, $f^{1}$ represents convolution layer with kernel size of $1\times1$. In Eq. \ref{p1} and Eq. \ref{p2}, $\mathbb{PS}(.)$ is the PixelShuffle with upsample scale factor of 2.\\
\textbf{Subspace Pyramid Fusion Module (SPFM)}: The final output for SPFM module $\hat{P}_{out}$ is obtained by concatenating RPP of all splits (Eq. \ref{pf}).

\begin{equation}\label{pf}
	\hat{P}_{out} = f_{1}( Cancat([\hat{P}_{\tilde{s},1},\hat{P}_{\tilde{s},2},...,\hat{P}_{\tilde{s},s}]))
\end{equation} 
\begin{figure*}[t]
	\centering
	\includegraphics[width=0.9\linewidth]{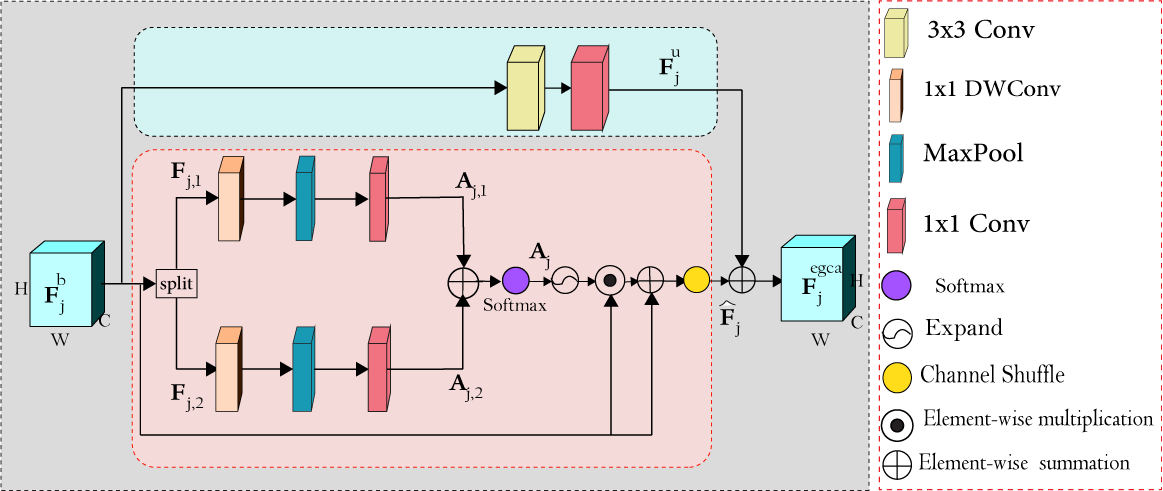}
	\caption{An overview of the proposed Global Context Aggregation module.
		It utilize channel split/shuffle to process features in each group.	}
	\label{Figure:egca}
\end{figure*}

\subsection{Efficient Shuffle Attention Module(ESAM)}
\label{sec:shuffle_attention}
The significance of local and global context information in semantic segmentation and various computer vision tasks has been acknowledged, even in the traditional machine learning era \cite{arbelaez2010contour}. The Efficient Shuffle Attention Module (ESAM) is introduced to amalgamate channel attention via the shuffle unit operator, thereby enhancing cross-feature interaction among sub-features. In this context, the input features from the backbone are represented as 
$F_{j}^{b}$, where j denotes the level in the encoder (refer to Figure \ref{spfnet}). Considering the distinct semantic content across different levels, ESAM is applied before upsampling to a common resolution using bilinear interpolation. Subsequently, features from all encoder scales and the final decoder-based subpixel convolution outputs are concatenated to formulate a multi-scale feature map. This process facilitates the fusion of information from both shallow and deeper layers. 
 
Given the $j^{th}$ feature $F_{j}^{b}\in \mathbb{R}^{H\times W\times C}$ from the backbone, where H and W are spatial dimensions and $C$ is the channel dimension. The output $F_{j}^{egca}$ of efficient shuffle attention module is computed using the following equation:\\
\begin{equation}\label{eq:egca}
{F}_{j}^{egca} = \widehat{F}_{j} \oplus F_{j}^{u}
\end{equation} 

Where $\oplus$ represents the element-wise summation, $\widehat{F}_{j}$ is the output of shuffle attention part, and $F_{j}^{u}$ represents the output feature of the upper part. At first, the input feature $F_{j}^{b}$ is used to process information into upper branch and the shuffle attention branch (see Figure. \ref{Figure:egca}). More details are given as follow:\\
\textbf{Part I}: Upper branch\\
in this branch $F_{j}^{b}$ is passed through depthwise separable convolutions, leading to feature map of $F_{j}^{u}\in \mathbb{R}^{H\times W\times C}$ as shown in Eq. \ref{fu_upper}.

\begin{equation}\label{fu_upper}  
	F_{j}^{u} = f_{DW}^{3}(f^{1}(F_{j}^{b}))
\end{equation}

where $f_{DW}^{3}(.)$ represents depthwise convolution with kernel $3\times3$.\\
\textbf{Part II}: Lower branch- shuffle attention\\
The second branch uses shuffle channel attention  to  generate sub-features that gradually capture multi-scale semantic responses. In this branch, the input $F_{j}^{b}$ is divided  into two groups $F_{j,1}\in \mathbb{R}^{H\times W\times C/2}$, and $F_{j,2}\in \mathbb{R}^{H\times W\times C/2}$ along the channel dimensions. 

 \begin{equation}\label{sa1}
	A_{j,1}  =  f_{PW}^{1}(\mathbb{MP}(f_{DW}^{1}(F_{j,1})))
\end{equation}

\begin{equation}\label{sa2}
	A_{j,2} = f_{PW}^{1}(\mathbb{MP}(f_{DW}^{1}(F_{j,2})))
\end{equation}

$A_{j,1}\in \mathbb{R}^{H\times W\times C}$ ,and $A_{j,1}\in \mathbb{R}^{H\times W\times C}$, have the same operations. Each of them starts with depth-wise convolution followed by Max Pooling and Point-wise. The two part are combined to form an attention map $A_{j}$ for each sub-group to captures the long-range dependencies. 

\begin{equation}\label{sa3}
	A_{j} = Softmax(A_{j,1}\oplus A_{j,2})
\end{equation}

The output $\widehat{F}_{j}$ is computed using Eq. \ref{2nd_branch}. 
\begin{equation}\label{2nd_branch}
\widehat{F}_{j} = Shuffle((A_{j}\odot F_{j}^{b})\oplus F_{j}^{b})
\end{equation} 

Where $\odot$ is the element-wise multiplication. $f_{DW}^{1}(.)$ represents a depth-with convolution with kernel size of $1\times1$, $\mathbb{MP}(.)$ is a Max Pooling  and Point-wise $f_{PW}^{1}(.)$.
 
 The obtained output undergoes normalization through the softmax function, as expressed in Equation \ref{sa3}. Subsequently, element-wise multiplication and summation operations are implemented within residual-like connections involving the input feature $A_{j}$. Additionally, we incorporate a channel shuffle operator, akin to ShuffleNet \cite{ma2018shufflenet}, to facilitate cross-group information communication. The resulting channel attention output, as delineated in Equation \ref{2nd_branch}, is then summated with the upper branch to compose the Efficient Shuffle Attention Module (ESAM), as outlined in Equation \ref{eq:egca}. Importantly, the final output dimensions of the ESAM module align with those of $F_{i}^b$, rendering ESAM a straightforward and adaptable component that can be seamlessly integrated into diverse stages of the encoder. This flexibility empowers the module to acquire multi-scale features with distinct semantic attributes. Consequently, the ESAM module proves conducive to effortless integration into contemporary semantic segmentation architectures and analogous tasks. The impact of the proposed efficient context aggregation module is further illustrated in Table \ref{tab:egca_stages}, where evaluations encompass floating-point operations per second (FLOPS), parameters, processing speed, and mean intersection over union.

\begin{table*}[hpt]
	\centering
	\caption{\MakeUppercase{The Per-class, class, and category IoU  evaluation on the Cityscapes Test set.  ”-” indicates the corresponding result is not reported by the methods}}
	\label{Table:Tab8}
	\vspace{1ex}
	\begin{center}
		\begin{adjustbox}{width=1\textwidth}
			\small
			\begin{tabular}{l|c|c|c|c|c|c|c|c|c|c|c|c|c|c|c|c|c|c|c|c}
				\hline 				Method&Road&Sidewalk&Building&Wall&Fence&Pole&Traffic light&Traffic sign&Vegetation&Terrain&Sky&Person&Rider&Car&Truck&Bus&Train&Motor&Bicyclist&mIoU\\
				\hline
				\hline
			CRF-RNN\cite{zheng2015conditional}&96.3&73.9&88.2&47.6&41.3&35.2&49.5&59.7&90.6&66.1&93.5&70.4&34.7&90.1&39.2&57.5&55.4&43.9&54.6&62.5\\
				FCN\cite{long2015fully}&97.4&78.4&89.2&34.9&44.2&47.4&60.1.5&65.0&91.4&69.3&93.9&77.1&51.4&92.6&35.3&48.6&46.5&51.6&66.8&65.3\\
			DeepLabv2\cite{chen2017deeplab}&-&-&-&-&-&-&-&-&-& -&-&-&-&-&-&-&-&-&-&70.4\\
			Dilation10\cite{yu2016multi}&97.6&79.2&89.9&37.3&47.6&53.2&58.6&65.2&91.8&69.4&93.7&78.9&55.0&93.3&45.5&53.4&47.7&52.2&66.0&67.1\\
				AGLNet\cite{zhou2020aglnet}&97.8& 80.1& 91.0&\textbf{51.3}&50.6&58.3&63.0&68.5&92.3&71.3&94.2&80.1&59.6&93.8&48.4&68.1&42.1&52.4& 67.8&70.1\\
				BiSeNetV2\slash BiSeNetV2\_L\cite{yu2021bisenet}&-&-&-&-&-&-&-&-&-& -&-&-&-&-&-&-&-&-&-&73.2\\
			LBN-AA\cite{dong2020real}&98.2&84.0&91.6&50.7&49.5&60.9&69.0&73.6&92.6& 70.3&94.4&83.0&65.7&94.9&62.0&70.9&53.3&62.5&71.8-&73.6\\
				\hline
			Our(SPFNet)&98.5&85.2&92.6&48.4&55.8&\textbf{67.0}&74.5&77.8&\textbf{93.4}&70.7&95.2&85.7&68.8&95.2&53.3&73.0&59.7&67.2&75.4&75.7\\
			Our(SPFNet)&98.4&84.9&92.2&47.6&54.1&65.9&73.6&76.2&93.0&66.4&94.8&85.3&68.5&94.6&47.6&53.9&28.8&65.7&75.0&71.9\\
			\hline
			\end{tabular}
		\end{adjustbox}
	\end{center}
\end{table*}  
\section{Experiments}
In this section, we conduct a comprehensive set of experiments to assess the efficacy of the proposed SPFNet. The evaluations are performed on the Camvid \cite{brostow2009semantic} and Cityscapes \cite{cordts2016cityscapes} datasets. The experimental results showcase SPFNet's state-of-the-art performance on the Camvid dataset and comparable results to other methods on the Cityscapes dataset. The subsequent sections detail the datasets and implementation specifics, followed by a series of ablation studies conducted on the Camvid and Cityscapes datasets to elucidate the impact of each component in SPFNet. Finally, we present a comparative analysis of our method with other state-of-the-art networks on both datasets.

\subsection{Experiments Settings}
\subsubsection{Camvid}
The Camvid dataset \cite{brostow2009semantic} comprises 376 training examples, 101 evaluation images, and 233 testing images. To ensure a fair comparison with other state-of-the-art models, we assessed our model across 11 classes, including building, sky, tree, car, and road. The 12th class was designated as an ignore class to accommodate unlabelled data. The dataset's small size and imbalanced label distribution pose significant challenges.
\subsubsection{Cityscapes}
The Cityscapes dataset \cite{cordts2016cityscapes} serves as an urban scene understanding benchmark, featuring 5000 high-resolution images of dimensions 2048 x 1024 pixels with fine annotations from various cities. The dataset is divided into 2975 images for training, 500 images for validation, and the remaining 1525 images for testing. The evaluation spans 19 semantic segmentation classes.

The subsequent sections delve into the ablation studies, exploring the impact of each constituent in SPFNet on the Camvid and Cityscapes datasets. The experimental design aims to comprehensively assess the model's performance across diverse scenarios. Finally, a comparative analysis is presented, juxtaposing SPFNet against other state-of-the-art networks on the two datasets.

\subsubsection{Implementation Details}
To implement our model, we use Adam optimizer \cite{kingma2014adam} with weight decay $5\times10^{-6}$, power 0.9, and the starting learning rate $lr_{init}$ for Camvid and Cityscapes datasets is set to $3\times10^{-4}$. The network optimization utilized the Poly learning rate policy, as described in Equation \ref{eq:scheduler}, a strategy consistent with prior works such as \cite{chen2017deeplab, zhao2017pyramid}. All experiments were conducted using the PyTorch framework \cite{pytorch} on an NVIDIA 3090 RTX GPU, with training extending over 150 epochs for Camvid and 500 epochs for Cityscapes. Notably, BatchNorm layers in SPFNet, excluding those in the encoder, were substituted with InplaceBN-Sync \cite{bulo2018place} to curtail memory usage, enhance speed, and facilitate FLOPS analysis, which was conducted using an Nvidia GTX1080 GPU. Consistent data augmentation was applied across all experiments, involving random scaling between 0.75 and 2.0, random horizontal flips, and random cropping of $512\times1024$ and $360\times480$ image patches for training Cityscapes and Camvid, respectively. Weighted cross-entropy loss was employed for training on both datasets.To assess the experimental results of our model, comprehensive comparisons were conducted on the Cityscapes validation and testing sets against current state-of-the-art segmentation models. Further detailed analyses are presented in subsequent sections.
\begin{equation}\label{eq:scheduler}
lr = lr_{init}\times (1-\frac{iter}{max\_iter})^{power}     
\end{equation}
Where ${max\_iter}$ is the  maximum number of iterations.

\subsection{Ablation Study}
In this subsection, a series of experiments is conducted to assess the individual contributions of components within the proposed SPFNet. Initially, a baseline network is established, employing an encoder-decoder architecture consisting of a ResNet-based encoder and a decoder incorporating subpixel convolution. This baseline serves as the foundation for the subsequent ablation study.

\subsubsection{Ablation for Efficient Shuffle Attention Module(ESAM) }
\label{subsec:Ablation for efficient global aggregation(EGCA)module}
To assess the impact of the Efficient Shuffle Attention Module (ESAM) at each stage, we seamlessly integrate our proposed ESAM into a baseline U-shaped network architecture, denoted as ResNet34H (encoder) and DSC (decoder). The experimental results, presented in Table \ref{tab:egca_stages}, encompass evaluations in terms of FLOPS, parameters, processing speed, and mean Intersection over Union (mIoU). This integration enables the model to acquire multi-level cross-feature communication and mitigate the semantic gap among features from different levels. Importantly, the Subspace Pyramid Fusion Module (SPFM) is not incorporated into the network at this juncture. The introduction of ESAM leads to a marginal increase in computational requirements, as evident from the FLOPS results. However, the corresponding improvement in segmentation accuracy, as demonstrated in Table \ref{tab:egca_stages}, substantiates the efficacy of ESAM. Notably, integrating ESAM in the second stage results in a slightly higher FLOPS compared to other stages. In the configuration utilizing ESAM across stages 2 to 5, the mIoU experiences an improvement of over 2.84\% compared to the baseline. It is noteworthy that ESAM incurs a reduction in processing speed by 5.7 FPS compared to the baseline..\\
\begin{table}[hpt]
	\caption{\MakeUppercase{Evaluation of ESAM effect at different stages on Cityscapes validation set.FLOPS, SPEED ARE ESTIMATED FOR AN INPUT SIZE OF 512,1024.}}
	\label{tab:ESAM_stages}
	\begin{center}
		\begin{adjustbox}{width=0.48\textwidth}
			\small
			\begin{tabular}{l|c|c|c|c|c|c|c|c}
				\hline
    	     Baseline&$Stage_{2}$&$Stage_{3}$&$Stage_{4}$&$Stage_{5}$&FLOPS (G)&Params( M)&Speed (FPS)&mIoU(\%)\\ 
				\hline
				\hline
			    \checkmark&&&&&268.6&37.7&19.3&72.16\\
				\checkmark&\checkmark&&&&280.3&37.8&16.2&72.8\\
				\checkmark&\checkmark& \checkmark&&&287.0&37.9&15.2&73.19\\
				\checkmark&\checkmark& \checkmark & \checkmark&&293.6&38.0&14.2&74.01\\
				\checkmark&\checkmark& \checkmark& \checkmark& \checkmark&300.2&38.7&13.6&75.0\\
				\hline
			\end{tabular}
		\end{adjustbox}
	\end{center}
\end{table}
\subsubsection{Ablation on Subspace Pyramid Fusion Module (SPFM)}
\begin{figure}
	\begin{subfigure}[b]{0.23\textwidth}
		\includegraphics[width=4cm,height=4cm]{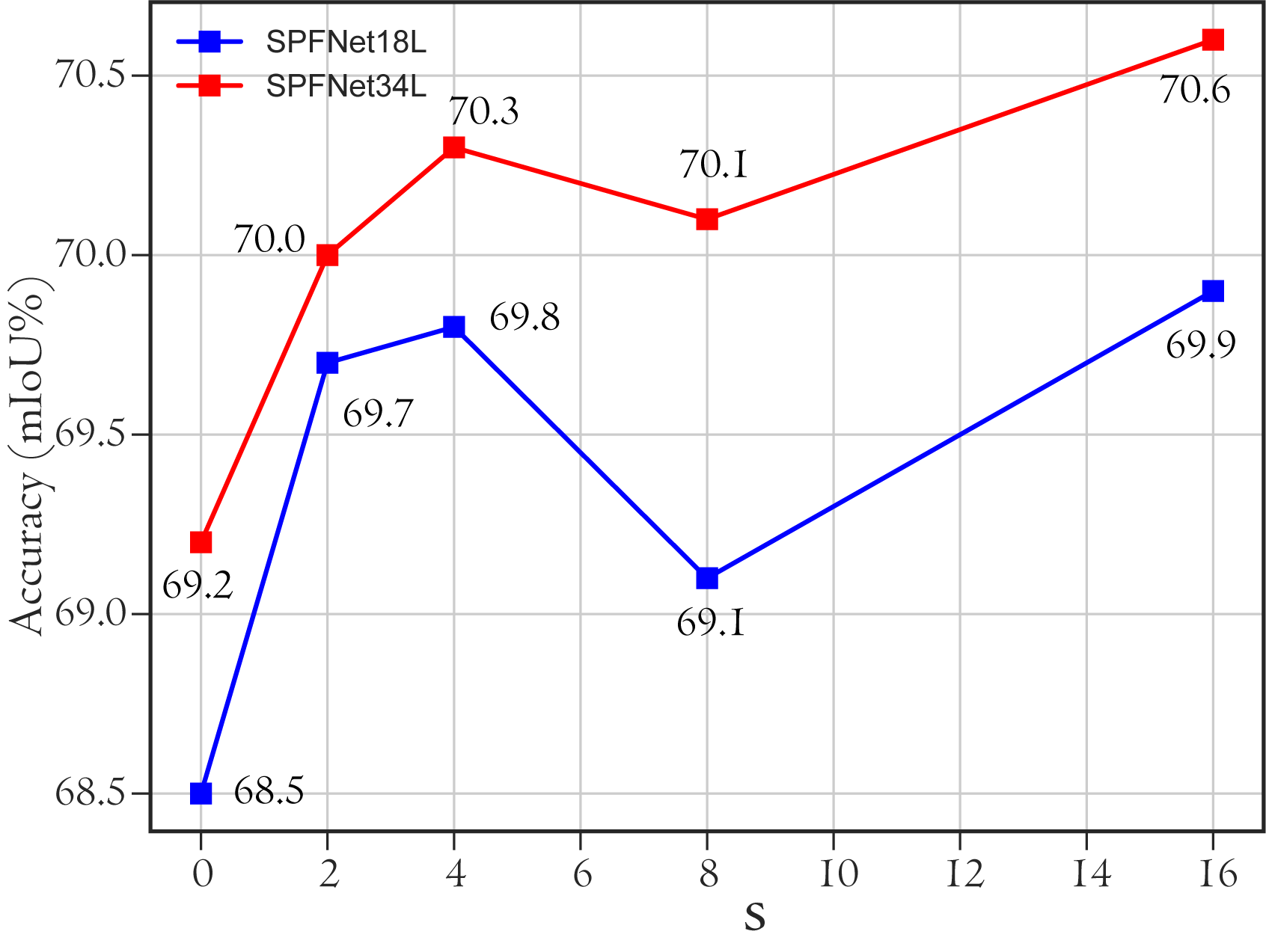}
		\caption{Hyperparameter vs accuracy.}
		\label{fig:hyperparameter_s_accuracy_camvid}
	\end{subfigure} \hspace{1.5mm}
	\begin{subfigure}[b]{0.23\textwidth}
		\includegraphics[width=4cm,height=4cm]{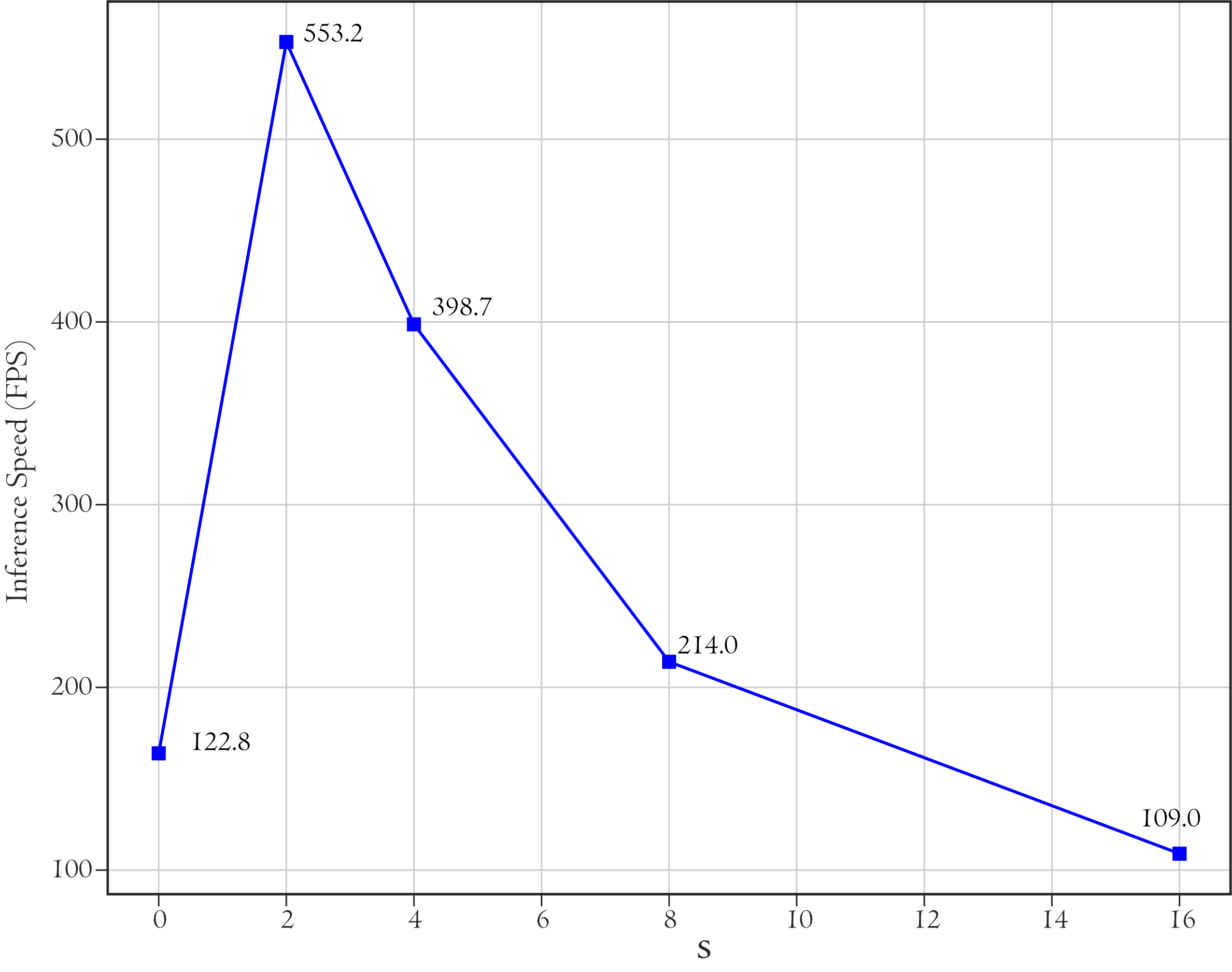}
		\caption{Hyperparameter vs FPS.}
		\label{fig:hyperparameter_s_speed_camvid}
	\end{subfigure}\hspace{1.5mm}
	\caption{(a) The accuracy obtained by the proposed SPFM with different values of the parameter s on the Camvid test set. (b) The inference speed with different s values.}
	\label{fig:hyperparameter_s_camvid}
\end{figure}

In our hyperparameter analysis on the Subspace Pyramid Fusion Module (SPFM) applied to the Camvid dataset, we systematically examine the impact of the hyperparameter s, representing the number of splits used in each Reduced Pyramid Pooling (RPP) module within the SPFM. The evaluations are conducted on both the Camvid and Cityscapes datasets, utilizing an Nvidia GTX1080 GPU. Specifically, we investigate four scenarios with ResNet18 (SPFNet18L) as the backbone and another four scenarios with ResNet34 (SPFNet23L), considering different values for s, namely$s={2,4,8,16}$. Initially, to establish a baseline, we replace the SPFM with a $1\times1$ Convolution layer, and the results for the Camvid dataset are reported in Table \ref{tab:compare_spfm_camvid}. When $s=2$, the SPFM fuses two Reduced Pyramid Fusion Modules to generate multi-scale features. As evident in Table \ref{tab:compare_spfm_camvid}, leveraging SPFM with varying numbers of splits demonstrates higher accuracy, measured in terms of mean Intersection over Union (mIoU), compared to the baseline. Moreover, models with larger values of s exhibit improved segmentation accuracy, indicating that higher s enhances the network's ability to extract complex multi-scale features for both SPFNet34L and SPFNet18L, albeit at the expense of speed (Frames Per Second, FPS). Table \ref{tab:compare_spfm_camvid} and Figure \ref{fig:hyperparameter_s_camvid}.\subref{fig:hyperparameter_s_speed_camvid} reveal a performance drop in SPFM when $s=8$ for both backbones in comparison to the baseline. Interestingly, utilizing 2 splits adds more parameters and Floating-Point Operations per Second (FLOPS) than larger s options, resulting in faster inference. We further compared SPFM with four splits in terms of FLOPS, number of parameters, and speed. As outlined in Table \ref{tab:compare_spfm_camvid}, a larger s introduces fewer FLOPS and parameters, yet it diminishes speed. Considering both efficiency and segmentation accuracy, our analysis concludes with the selection of $s=4$ for SPFM.
\begin{table}
	\centering
	\caption{\MakeUppercase{The Evaluation and analysis of the Hyperparameter s on the SPFM Module. The input size for the baseline on the Camvid test set is $360\times840$.the input size for SPFM is $512\times12\times15$.}}
	\label{tab:compare_spfm_camvid}
	\vspace{1ex}
	\begin{adjustbox}{width=0.48\textwidth}
	\begin{tabular}{l|cc|cc|cc|c}
	\hline
	\multicolumn{8}{c}{Baseline SPFNet-34L(ResNet34 as backbone)}\\
		\hline 
		\multirow{2}{*}{Network}&\multicolumn{2}{c|}{FLOPS (G)} & \multicolumn{2}{c|}{Params (M)}&\multicolumn{2}{c|}{Speed (FPS)} & \multirow{2}{*}{IoU (\%)} \\\cline{2-7}
		\multicolumn{1}{c|}{}&Overall&SPFM& Overall& SPFM & Overall & SPFM& \\ 
		\hline\hline
		Baseline   &22.8 &   &37.7&  &122.8&  & 69.2\\
		SPFM (s=2) &22.8 &3.0&37.7&6.7 &122.8&553.2&70.0\\
		SPFM (s=4) &22.8 &1.6&37.7&3.4&122.8&398.7&70.3\\
		SPFM (s=8) &22.8 &0.896&37.7&1.9&122.8&214.0&70.1\\
		SPFM (s=16)&22.8 &0.544&37.7&1.0&122.8&109.0&70.6\\
		\hline 
		\multicolumn{8}{c}{Baseline SPFNet-18L(ResNet18+DSC Module)}\\
		\hline
		Baseline   &16.4&   &27.6&  &163.9&&68.5 \\
		SPFM (s=2) &16.4&3.0&27.6&6.7&163.9&553.2&69.7\\
		SPFM (s=4) &16.4&1.6&27.6&3.4&163.9&398.7&69.8\\
		SPFM (s=8) &16.4&0.896&27.6&1.9&163.9&214.0&69.1 \\
		SPFM (s=16)&16.4&0.544&27.6&1.0&163.9&109.0&69.9\\
		\hline 
	\end{tabular}
	\end{adjustbox}
\end{table}

\begin{figure}
	\begin{subfigure}[b]{0.23\textwidth}
		\includegraphics[width=4cm,height=4cm]{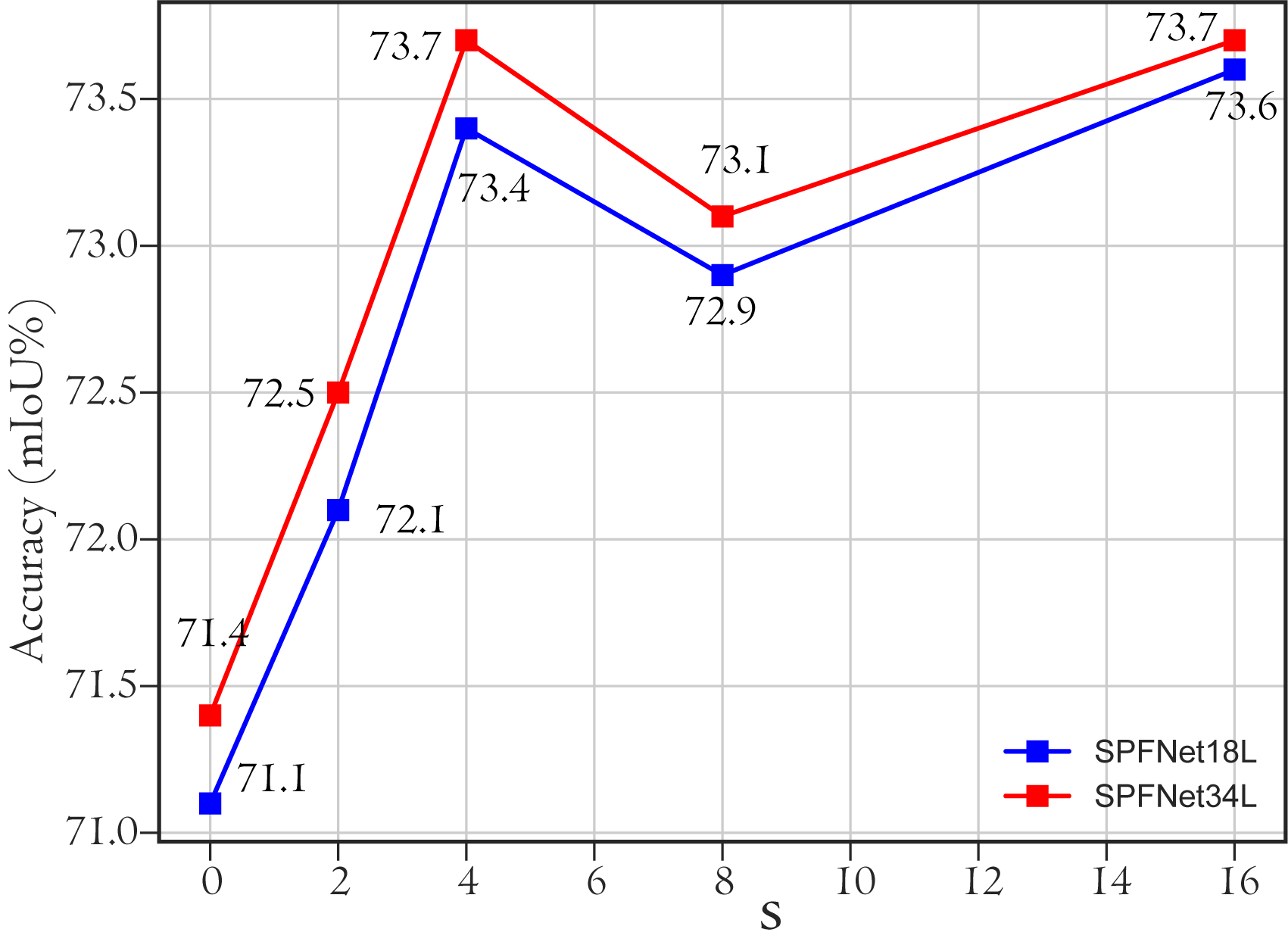}
		\caption{Hyperparameter vs accuracy.}
		\label{fig:hyperparameter_s_accuracy_cityscapes}
	\end{subfigure} \hspace{1.5mm}
	\begin{subfigure}[b]{0.23\textwidth}
		\includegraphics[width=4cm,height=4cm]{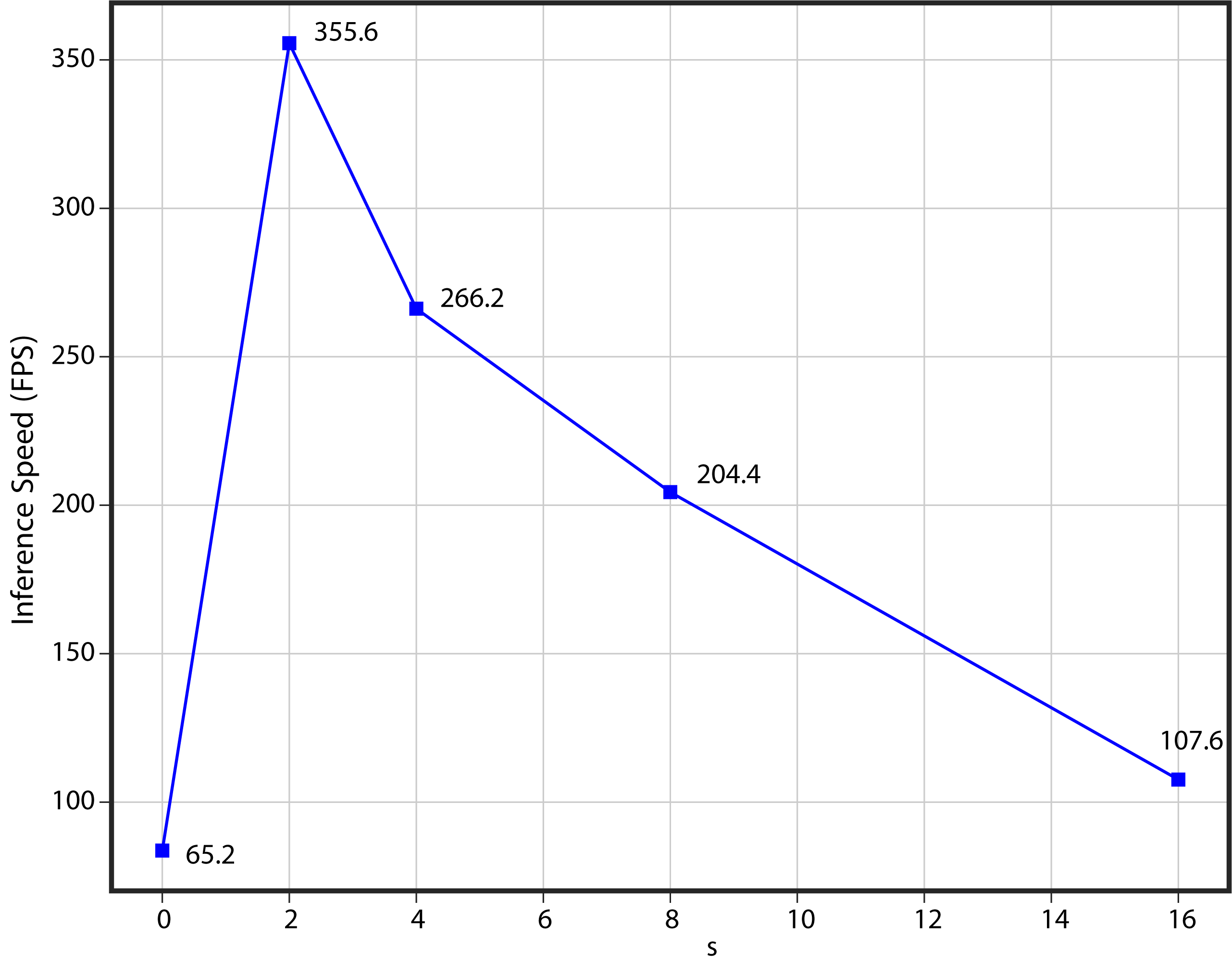}
		\caption{Hyperparameter vs FPS.}
		\label{fig:hyperparameter_s_speed_cityscapes}
	\end{subfigure}\hspace{1.5mm}
	\caption{(a) The accuracy obtained by the proposed SPFM with different values of the parameter s on the Cityscapes validation set. (b) The inference speed with different s values.}
	\label{fig:hyperparameter_s_cityscapes}
\end{figure}

Hyperparameter analysis on SPFM Module on the Cityscapes dataset:
We conducted similar analysis to examine the hyperparameter s of the SPFM on Cityscapes dataset in terms of the efficiency and accuracy. For this analysis we run the experiments for 150 epochs. From Table. \ref{tab:compare_spfm_cityscapes} and Figure. \ref{fig:hyperparameter_s_cityscapes} Experimental results and evaluation shows that the hyperparameter s follows the same patterns as on Camvid dataset.

\begin{table}
	\centering
	\caption{\MakeUppercase{The Evaluation and analysis of the Hyperparameter s on the SPFM Module.The input size for the baseline on the Cityscapes test set is $512\times1024$.input for the SPFM is $512\times16\times32$.}}
	\label{tab:compare_spfm_cityscapes}
	\vspace{1ex}
	\begin{adjustbox}{width=0.48\textwidth}
	\begin{tabular}{l|cc|cc|cc|c}
	\hline
	\multicolumn{8}{c}{Baseline SPFNet-34L(ResNet34 as backbone)}\\
		\hline 
		\multirow{2}{*}{Network}&\multicolumn{2}{c|}{FLOPS (G)} & \multicolumn{2}{c|}{Params (M)}&\multicolumn{2}{c|}{Speed (FPS)} & \multirow{2}{*}{mIoU (\%)} \\\cline{2-7}
		\multicolumn{1}{c|}{}&Overall&SPFM& Overall& SPFM & Overall & SPFM& \\ 
		\hline\hline
		Baseline   &68.1 &   &37.7&  &65.2&  & 71.4\\
		SPFM (s=2) &68.1 &8.6&37.7&6.7 &65.2&355.6&72.5\\
		SPFM (s=4) &68.1 &4.6&37.7&3.4&65.2&266.2&73.7\\
		SPFM (s=8) &68.1 &2.5&37.7&1.9&65.2&204.4&73.1\\
		SPFM (s=16)&68.1 &1.5&37.7&1.0&65.2&107.6&73.7\\
		\hline 
		\multicolumn{8}{c}{Baseline SPFNet-18L(ResNet18+DSC Module)}\\
		\hline
		Baseline   &48.7&   &27.6&  &83.7&&71.1 \\
		SPFM (s=2) &48.7&8.6&27.6&6.7&83.7&355.6&72.1\\
		SPFM (s=4) &48.7&4.6&27.6&3.4&83.7&266.2&73.4\\
		SPFM (s=8) &48.7&2.56&27.6&1.9&83.7&204.4&72.9\\
		SPFM (s=16)&48.7&1.5&27.6&1.0&83.7&109.0&73.6\\
		\hline 
	\end{tabular}
	\end{adjustbox}
\end{table}
\noindent

\subsubsection{Ablation of SPFM and Other Multi-scale Modules}
To comprehensively investigate the impact of various modules within our proposed network architecture, we initially configured a modified ResNet34 by abstaining from downsampling in the second stage of the encoder. Subsequently, we integrated the Dilated Spatial Convolution (DSC) module and the Efficient Shuffle Attention Module (ESAM) module as baselines to assess different multi-scale feature extractor modules. These ablation experiments were conducted over 500 epochs.

In particular, we performed a series of experiments to compare our proposed Subspace Pyramid Fusion Module (SPFM) with other multi-scale pyramid modules, namely Atrous Spatial Pyramid Pooling (ASPP) from Deeplab \cite{chen2017deeplab}, Vortex Pooling \cite{xie2018vortex}, and DenseASPP \cite{yang2018denseaspp}. In these experiments, we opted for a somewhat lighter backbone, employing the aforementioned ResNet18 and a modified ResNet34. The results, summarized in Table \ref{tab:mult_scale_spfm}, reveal that all the multi-scale modules enhance the baseline. Notably, the baseline model incorporating our proposed SPFM outperforms those with ASPP (75.64\% mIoU), Vortex Pooling (75.72\% mIoU), and DenseASPP (75.21\% mIoU), achieving the highest accuracy with a 78.04 mIoU. Moreover, the SPFM variant requires fewer parameters and exhibits a speed profile closer to ASPP. Additionally, SPFM incurs lower computational cost, measured in terms of parameters and Floating-Point Operations per Second (FLOPS), compared to both DenseASPP and Vortex Pooling. This superiority could be attributed to the SPFM module's efficacy in extracting complex multi-scale semantic information more efficiently. SPFM concurrently learns multiple Reduced Pyramid Pooling (RPP) blocks, utilizing small dilation rates to generate feature maps of different resolutions, thereby minimizing noise and enhancing its overall performance.
\begin{table}
    \centering
	\caption{\MakeUppercase{Performance comparison of SPFM with state-of-the-art multi-scale extractor methods such as ASPP, DenseASPP, and Vortex Pooling on the Cityscapes Validation. FLOPS, SPEED ARE ESTIMATED FOR AN INPUT SIZE OF 512,1024. Under the SPFNet34H setting.}}
	\label{tab:mult_scale_spfm}
	\vspace{1ex}
	\begin{adjustbox}{width=0.48\textwidth}
			\small
		\begin{tabular}{l|c|c|c|c}
			\hline
			Method&FLOPS&Params&FPS&mIoU(\%)\\ 
			\hline
			\hline
			ResNet34+DSC baseline&268.6&37.7&19.3&72.16\\
			ResNet34+DSC+ESAM&300.0&38.7&13.6&75.0\\
			ResNet34+DSC+SPFM&286.6 &40.8&16.3&75.8\\
			ResNet34+DSC+ESAM+DenseASPP&380.7&77.9&8.7&75.21\\
			ResNet34+DSC+ESAM+ASPP&304.8&41.0&12.3&75.64\\
			ResNet34+DSC+ESAM+Vortex Pooling& 322.0&49.4&11.1&75.72\\
		    ResNet34+DSC+ESAM+SPFM& 317.0&41.8&12.4&78.04\\
			\hline
		\end{tabular}
	\end{adjustbox}
\end{table}
\subsection{Results on Cityscapes Dataset}
\begin{figure*}
	\centering
	\begin{subfigure}{0.24\textwidth}
		\centering
		\includegraphics[width=4.5cm,height=2.5cm]{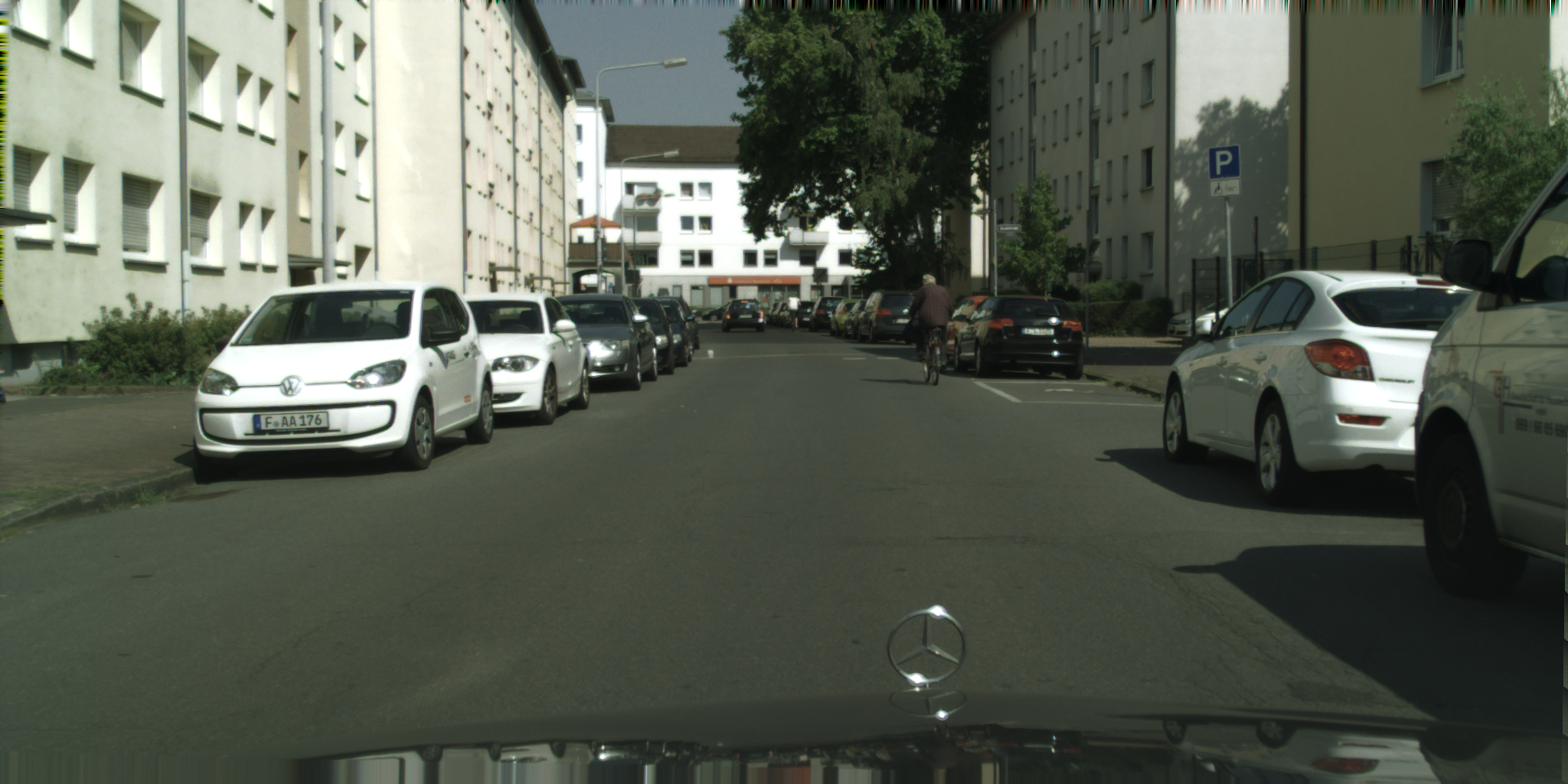}
	\end{subfigure}
	\begin{subfigure}{0.24\textwidth}
		\centering
		\includegraphics[width=4.8cm,height=2.5cm]{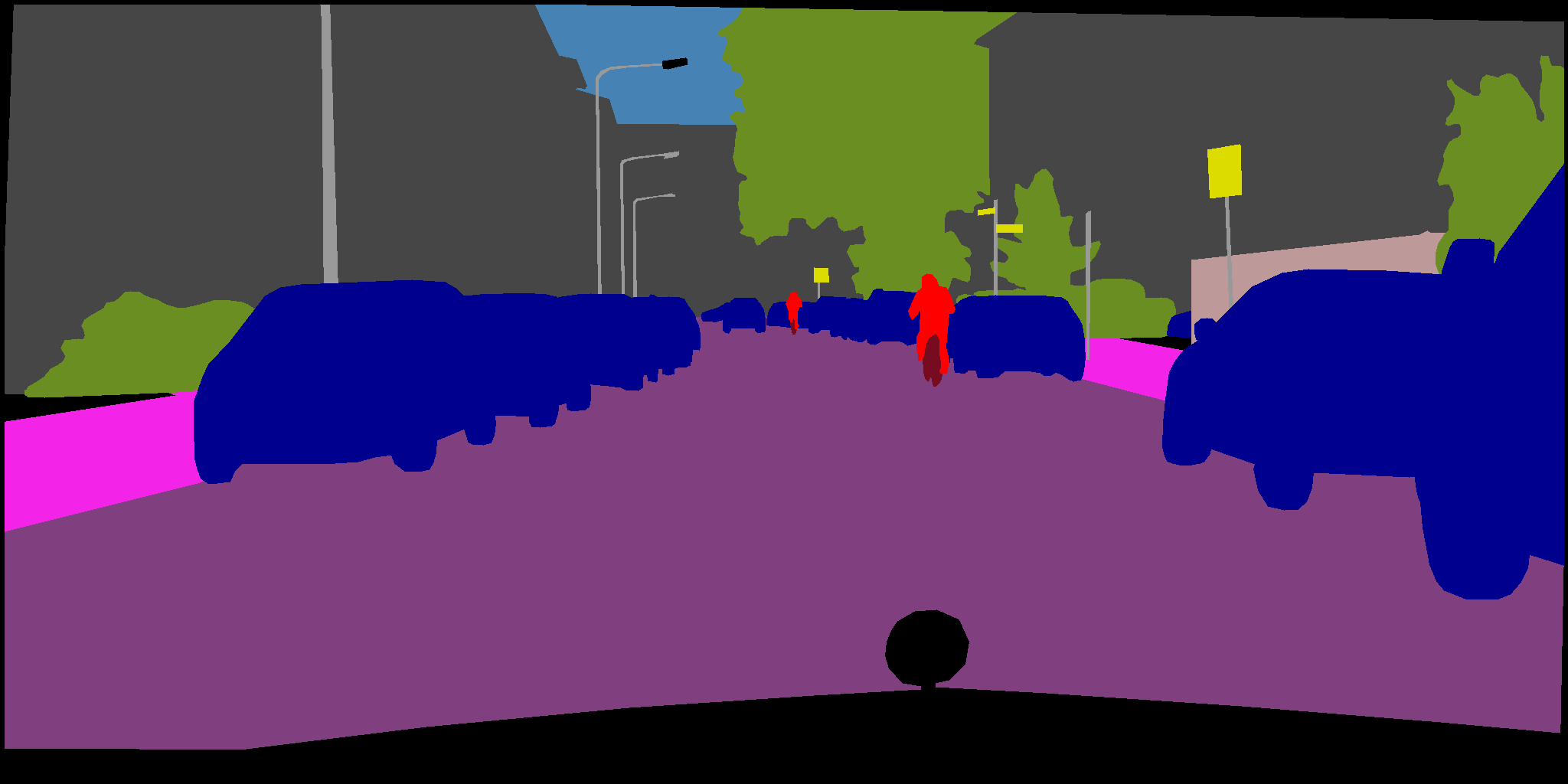}
	\end{subfigure}
	\begin{subfigure}{0.24\textwidth}
		\centering
		\includegraphics[width=4.8cm,height=2.5cm]{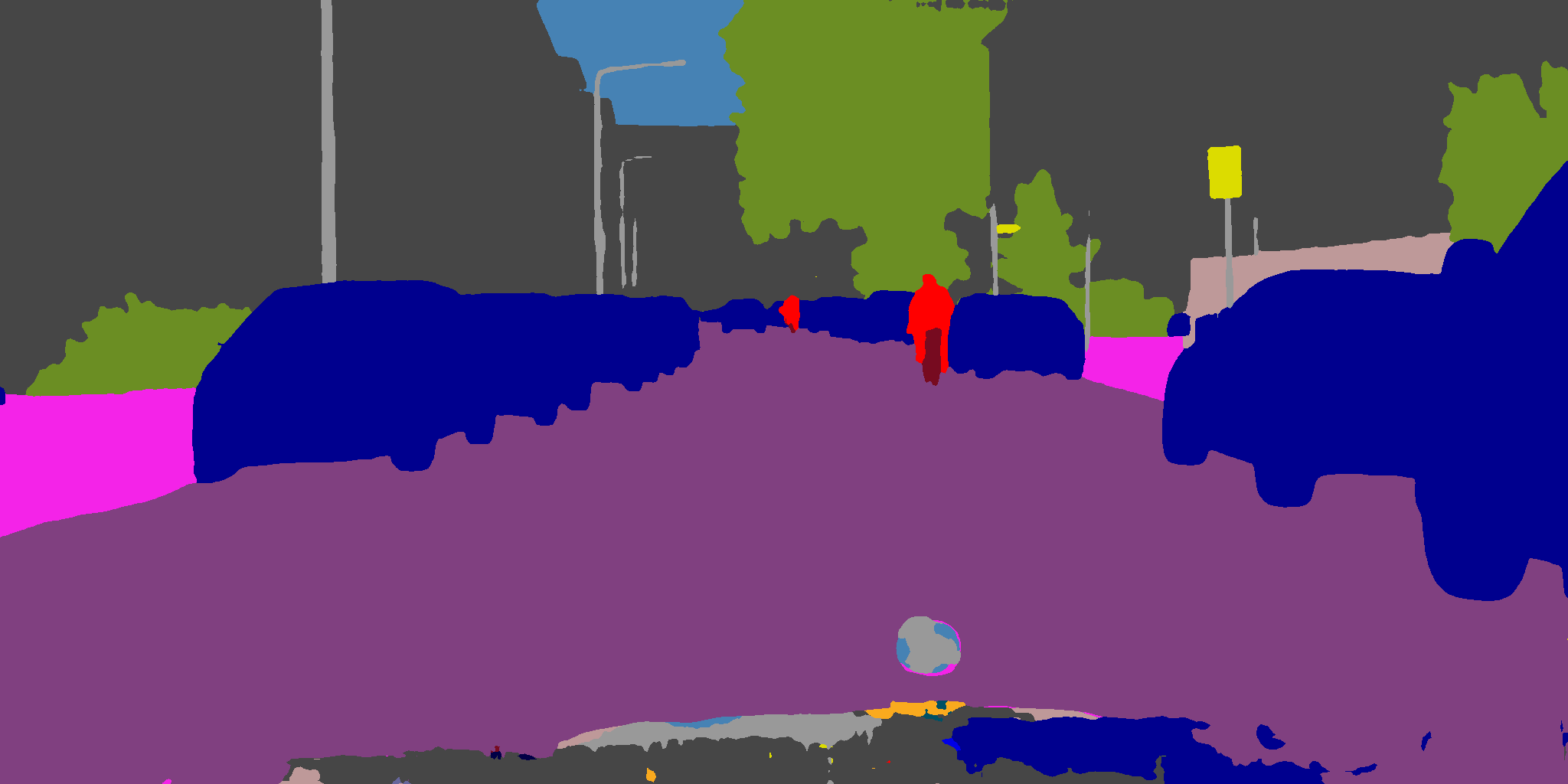}
	\end{subfigure}
	\\
	\begin{subfigure}{0.24\textwidth}
		\centering
		\includegraphics[width=4.8cm,height=2.5cm]{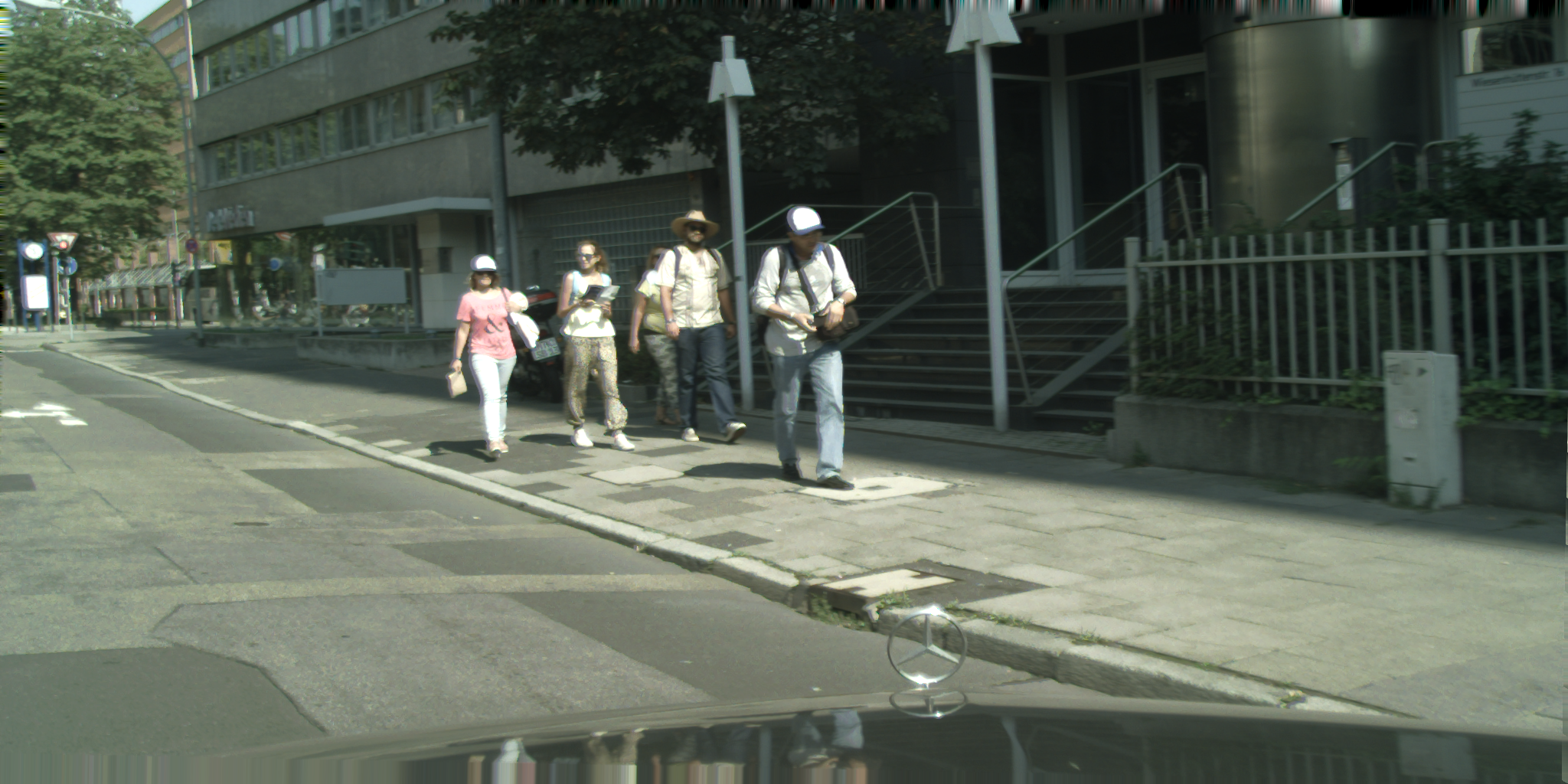}
	\end{subfigure}
	\begin{subfigure}{0.24\textwidth}
		\centering
		\includegraphics[width=4.8cm,height=2.5cm]{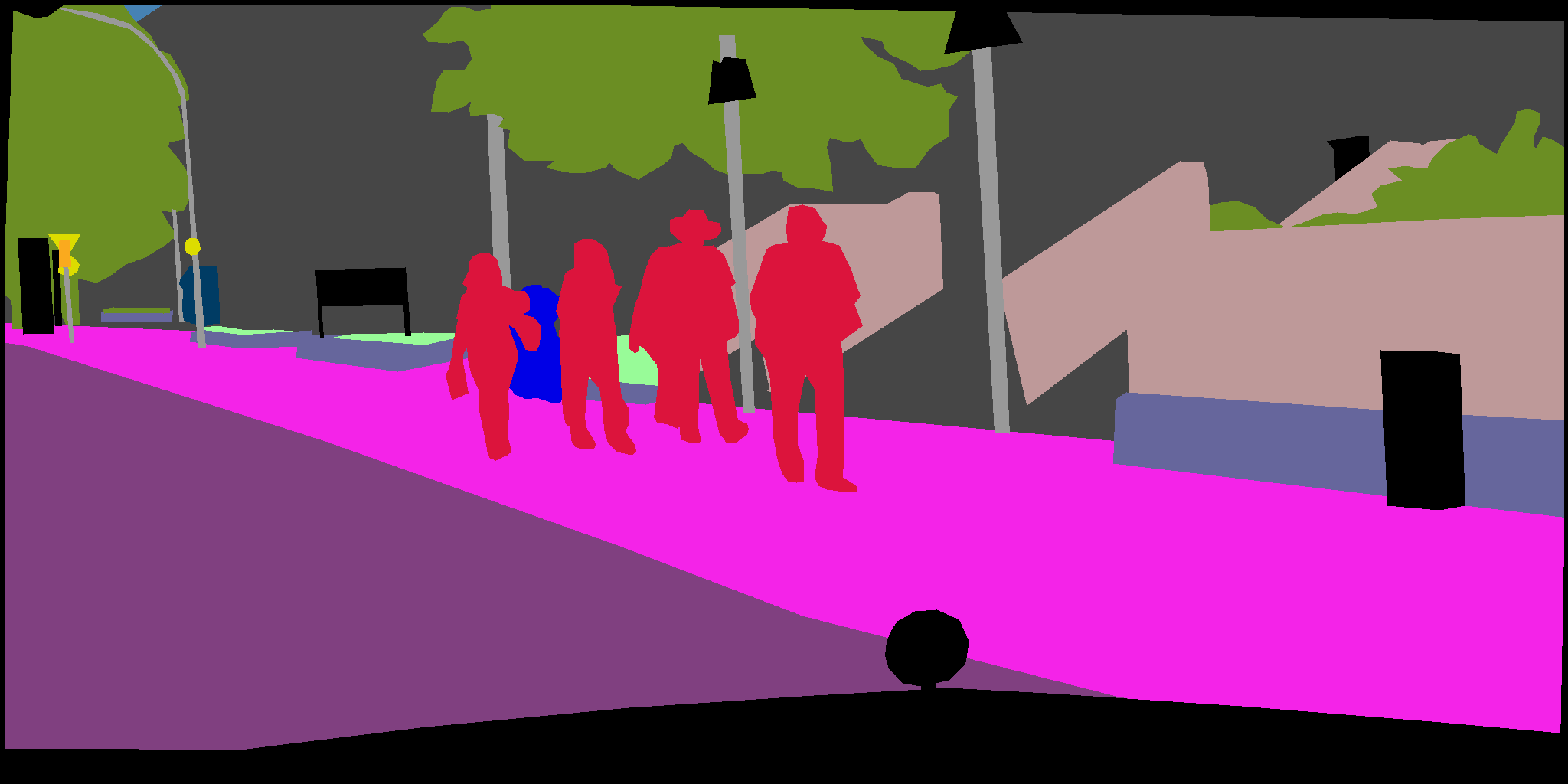}
	\end{subfigure}
	\begin{subfigure}{0.24\textwidth}
		\centering
		\includegraphics[width=4.8cm,height=2.5cm]{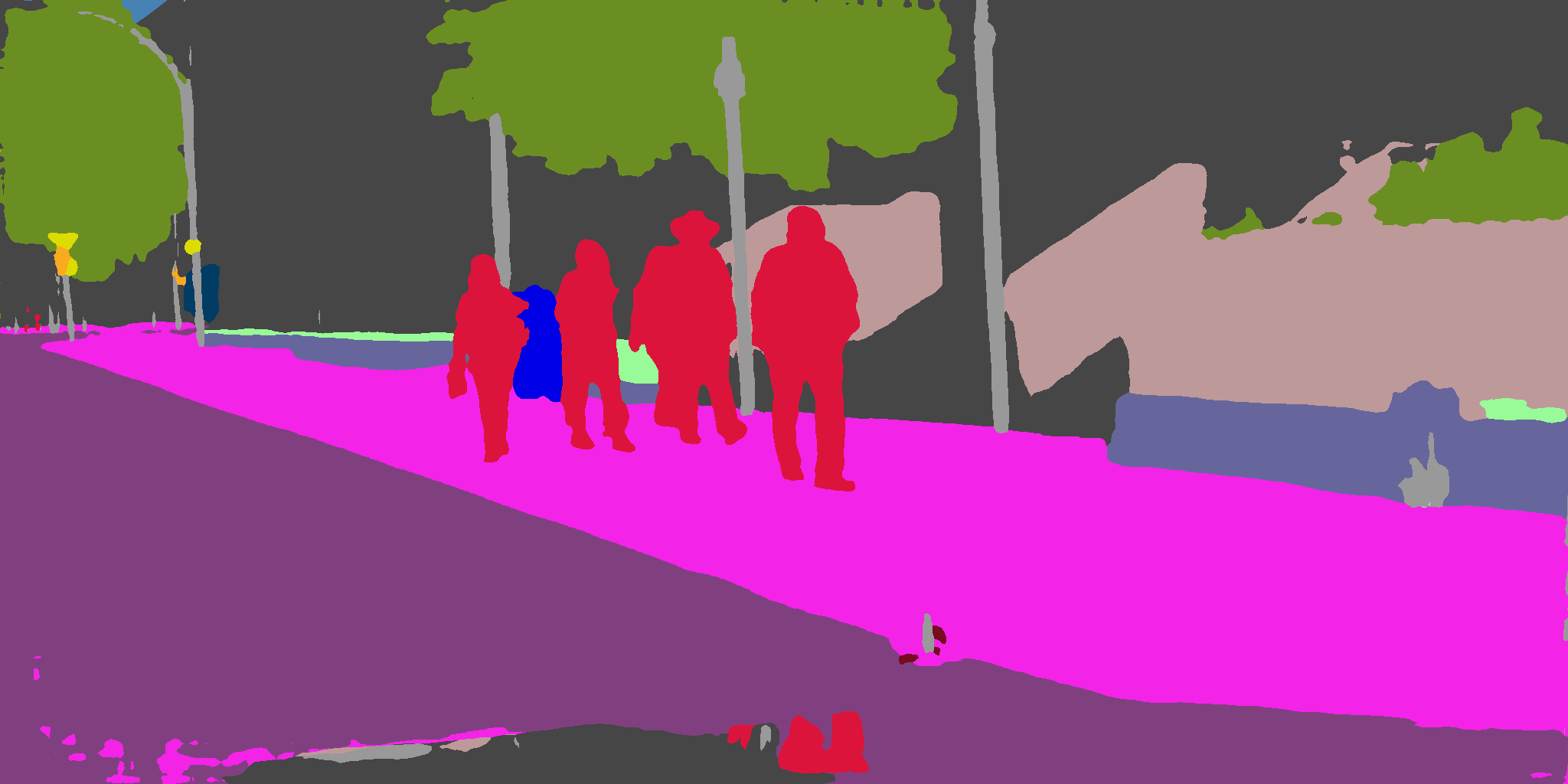}
	\end{subfigure}
	\\
	\centering
	\begin{subfigure}{0.24\textwidth}
		\centering
		\includegraphics[width=4.8cm,height=2.5cm]{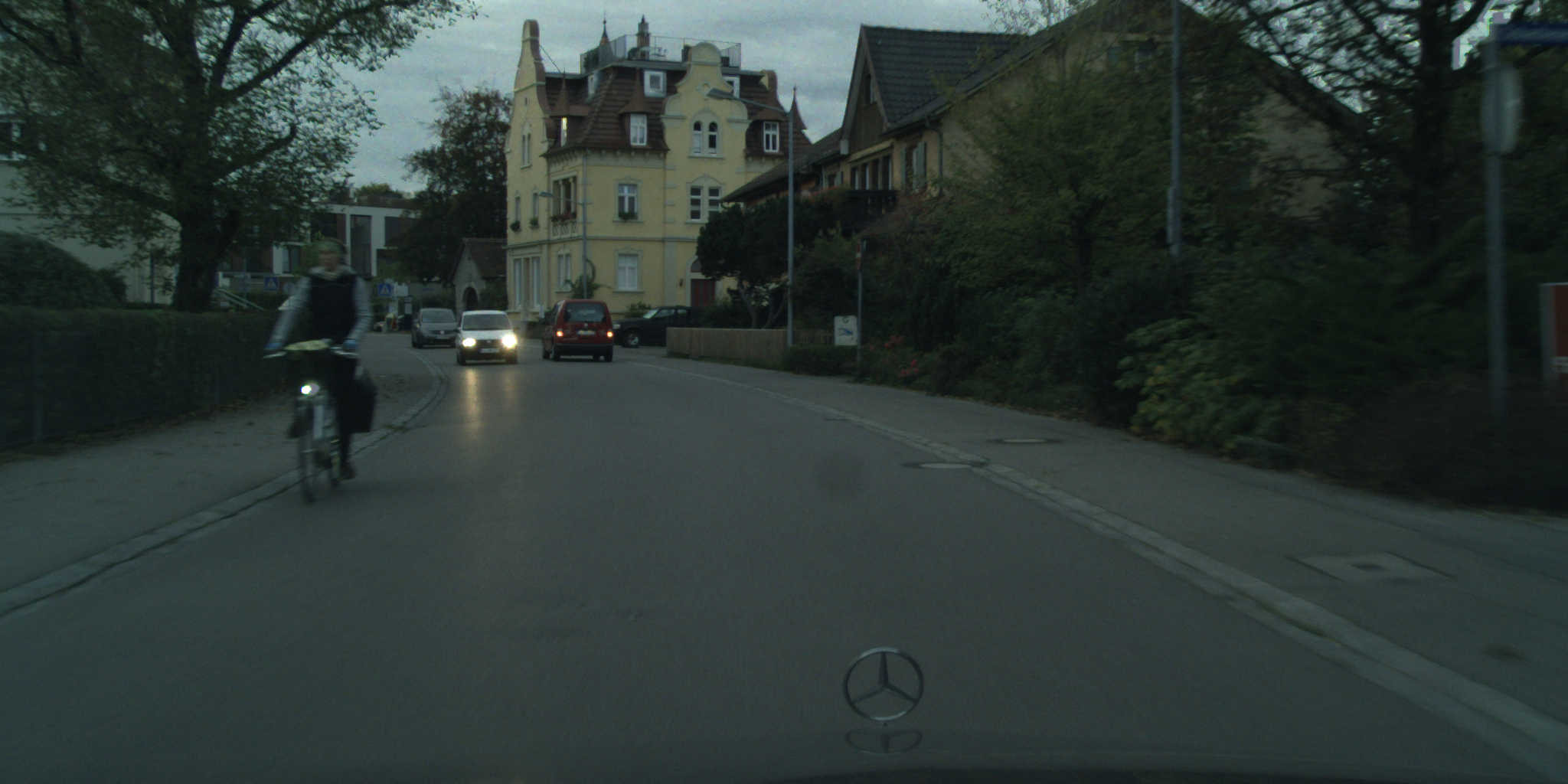}
	\end{subfigure}
	\begin{subfigure}{0.24\textwidth}
		\centering
		\includegraphics[width=4.8cm,height=2.5cm]{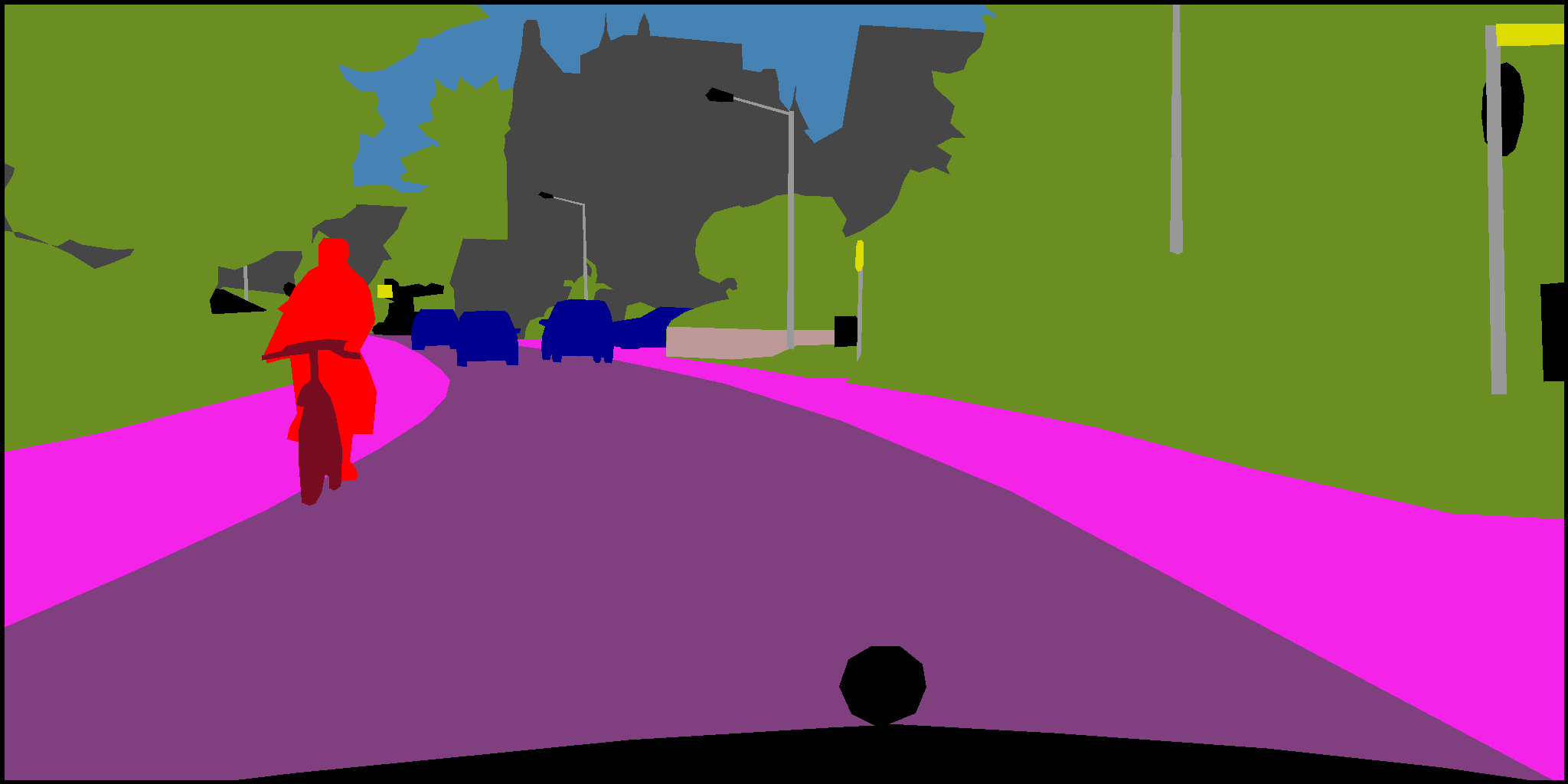}
	\end{subfigure}
	\begin{subfigure}{0.24\textwidth}
		\centering
		\includegraphics[width=4.8cm,height=2.5cm]{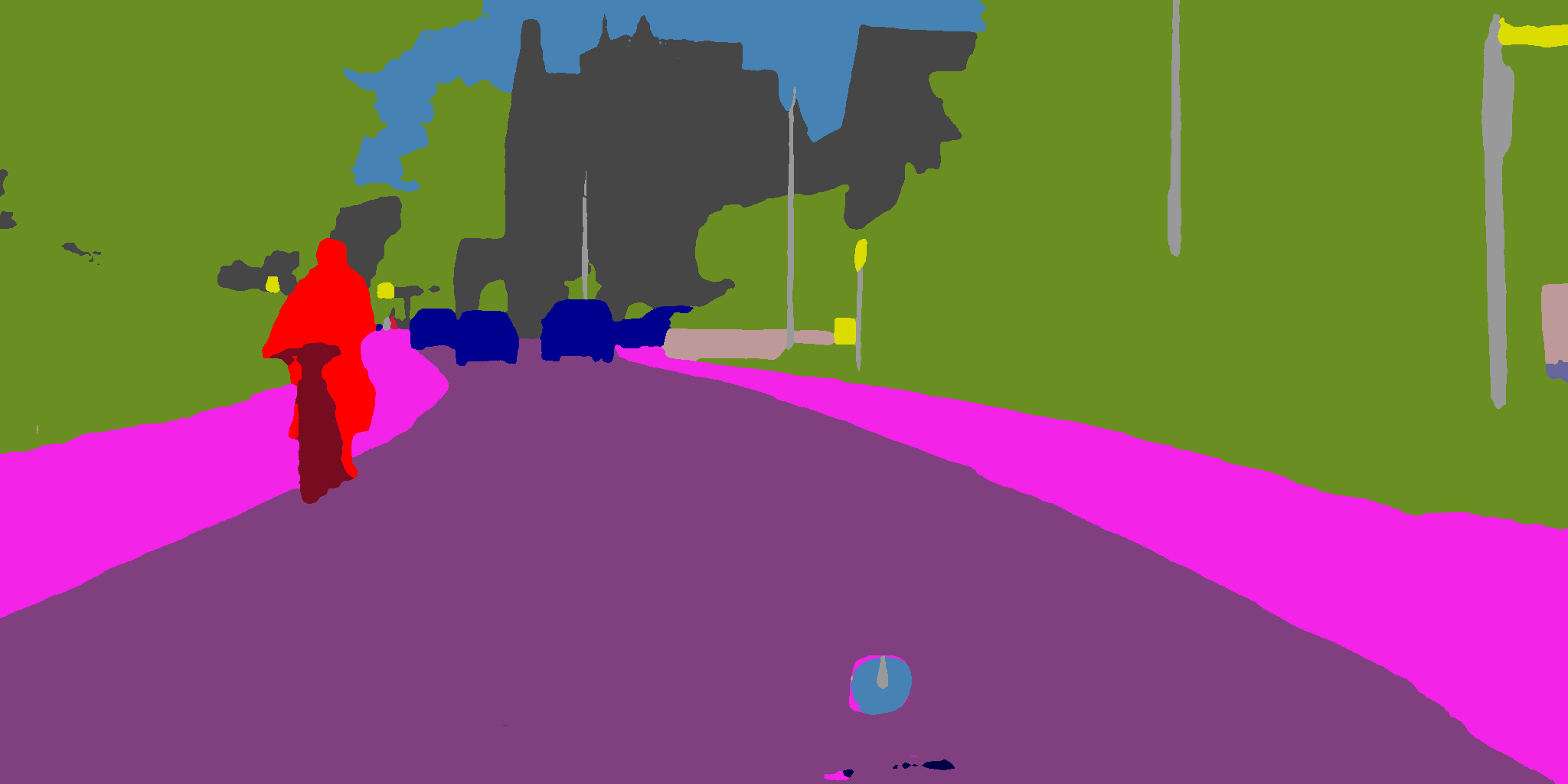}
	\end{subfigure}
	\\
	\centering
	\begin{subfigure}{0.24\textwidth}
		\centering
		\includegraphics[width=4.8cm,height=2.5cm]{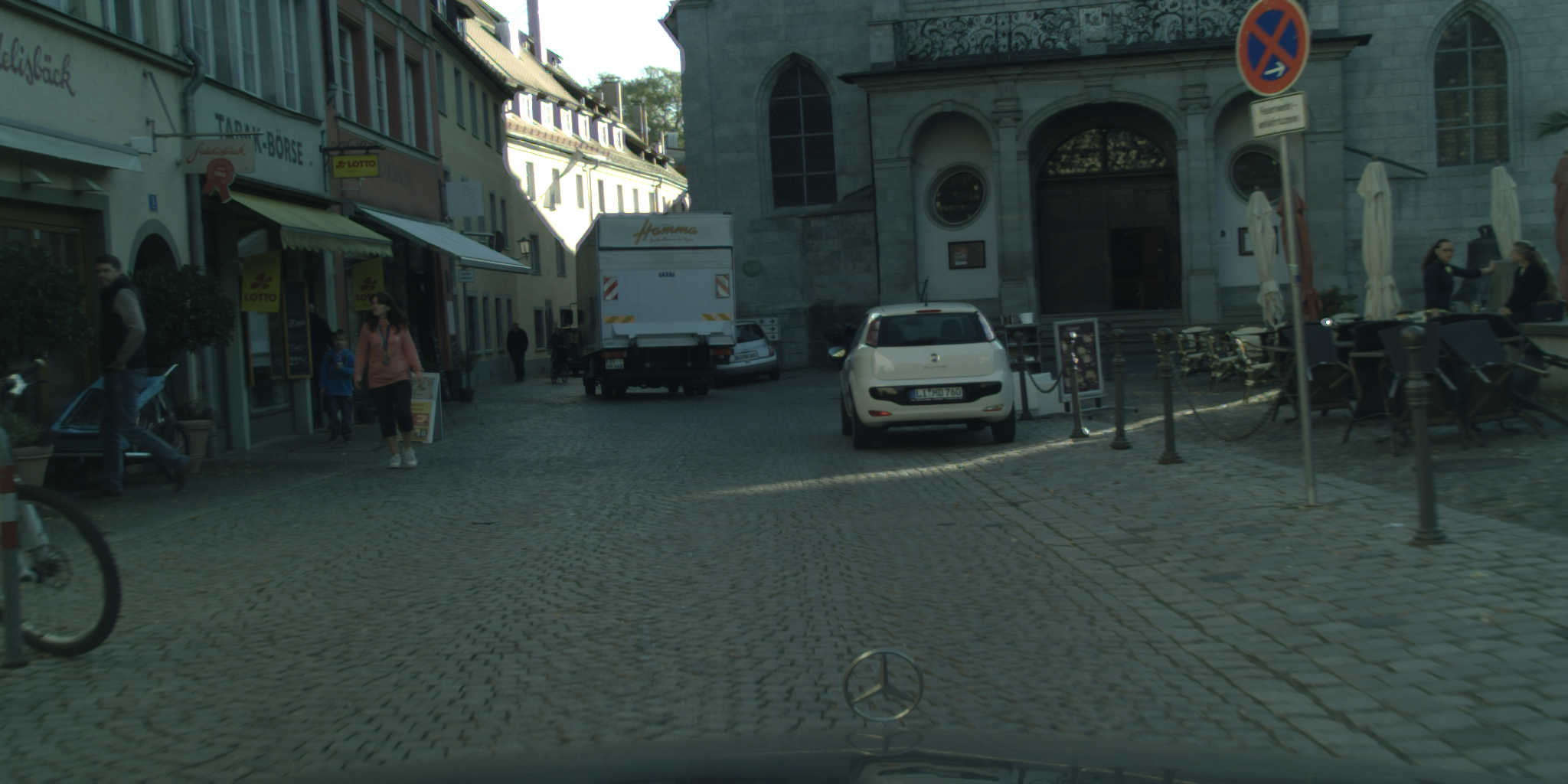}
	\end{subfigure}
	\begin{subfigure}{0.24\textwidth}
		\centering
		\includegraphics[width=4.8cm,height=2.5cm]{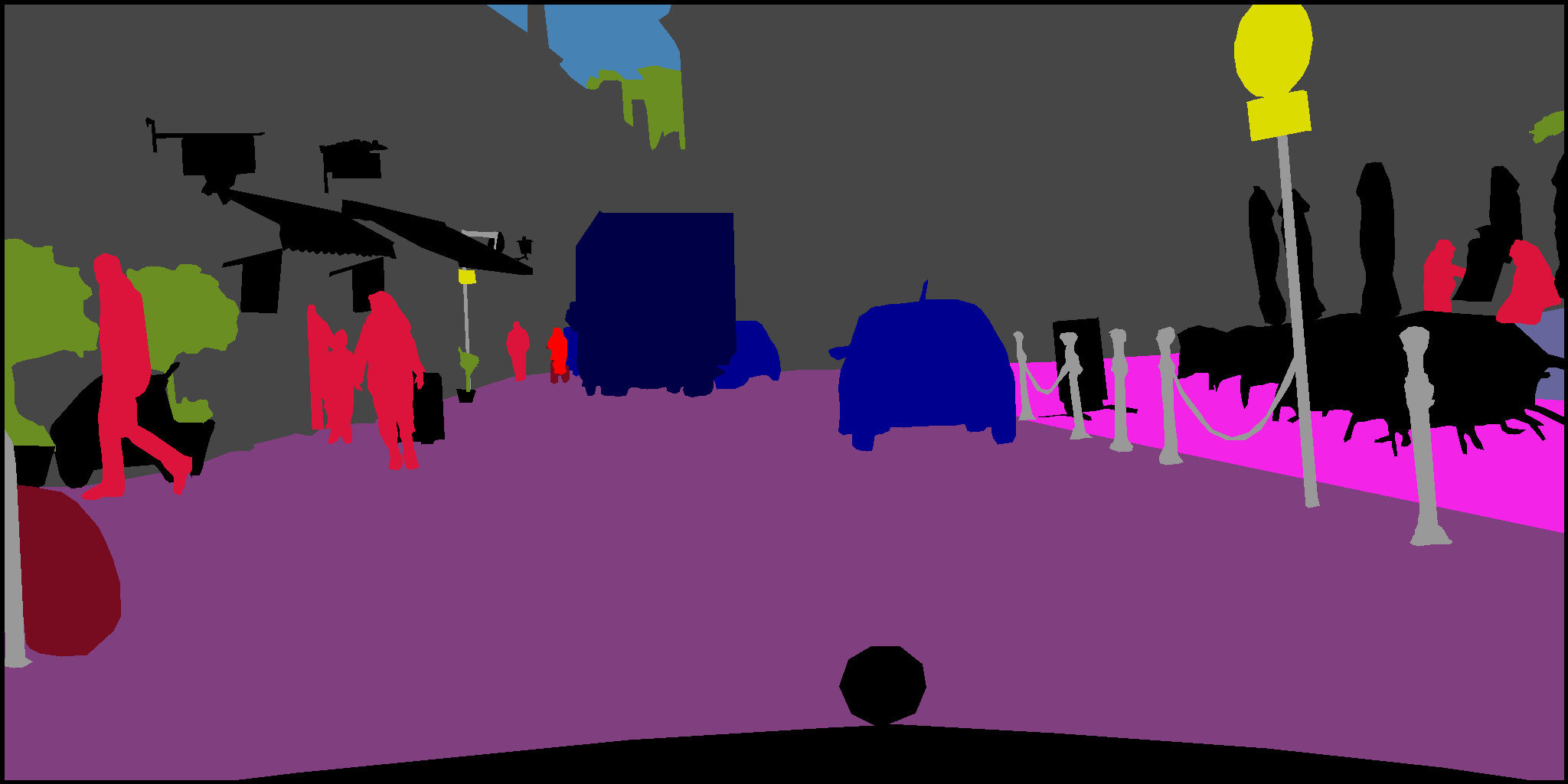}
	\end{subfigure}
	\begin{subfigure}{0.24\textwidth}
		\centering
		\includegraphics[width=4.8cm,height=2.5cm]{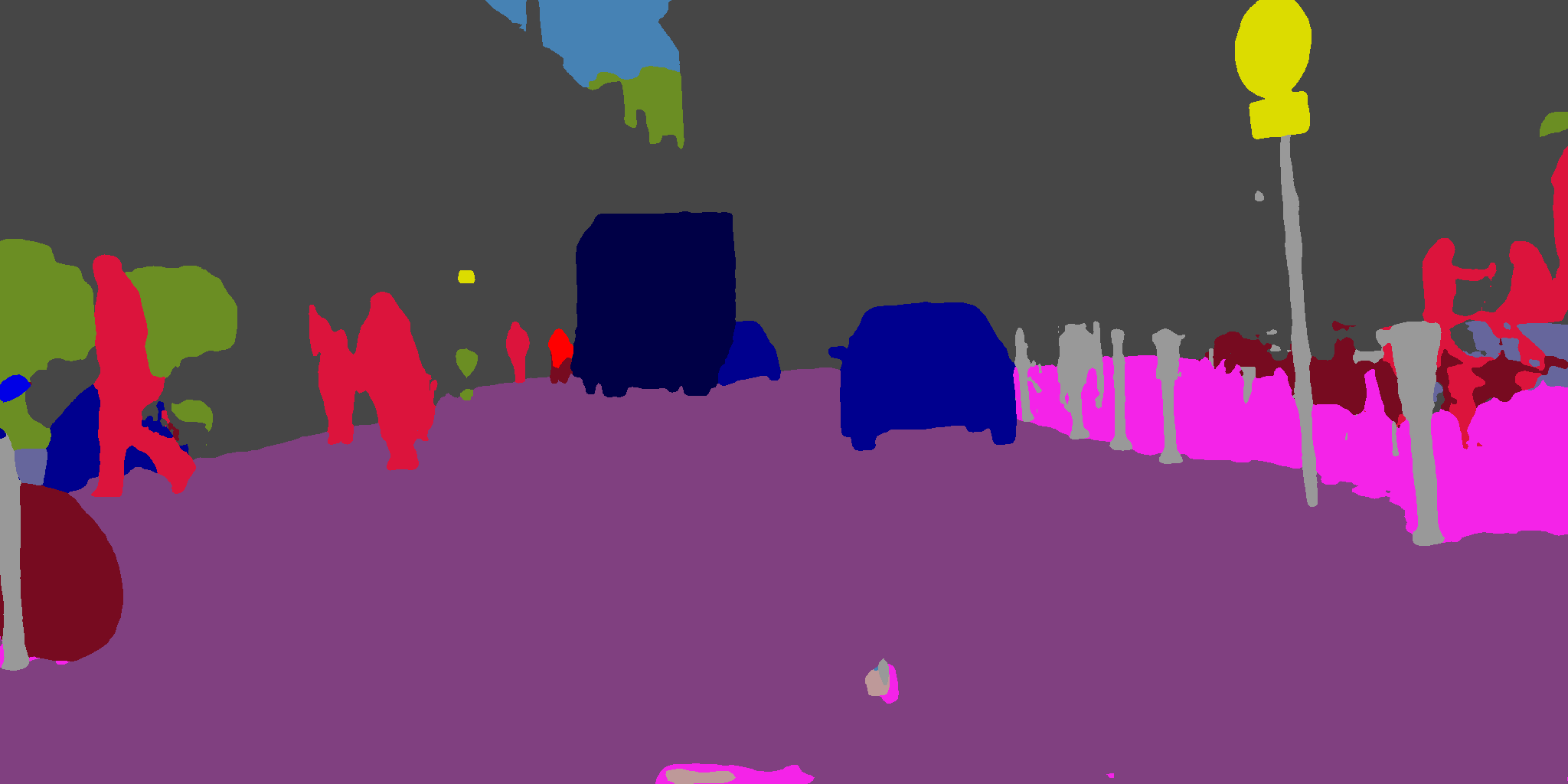}
	\end{subfigure}
	\\
	\centering
	\begin{subfigure}{0.24\textwidth}
		\centering
		\includegraphics[width=4.8cm,height=2.5cm]{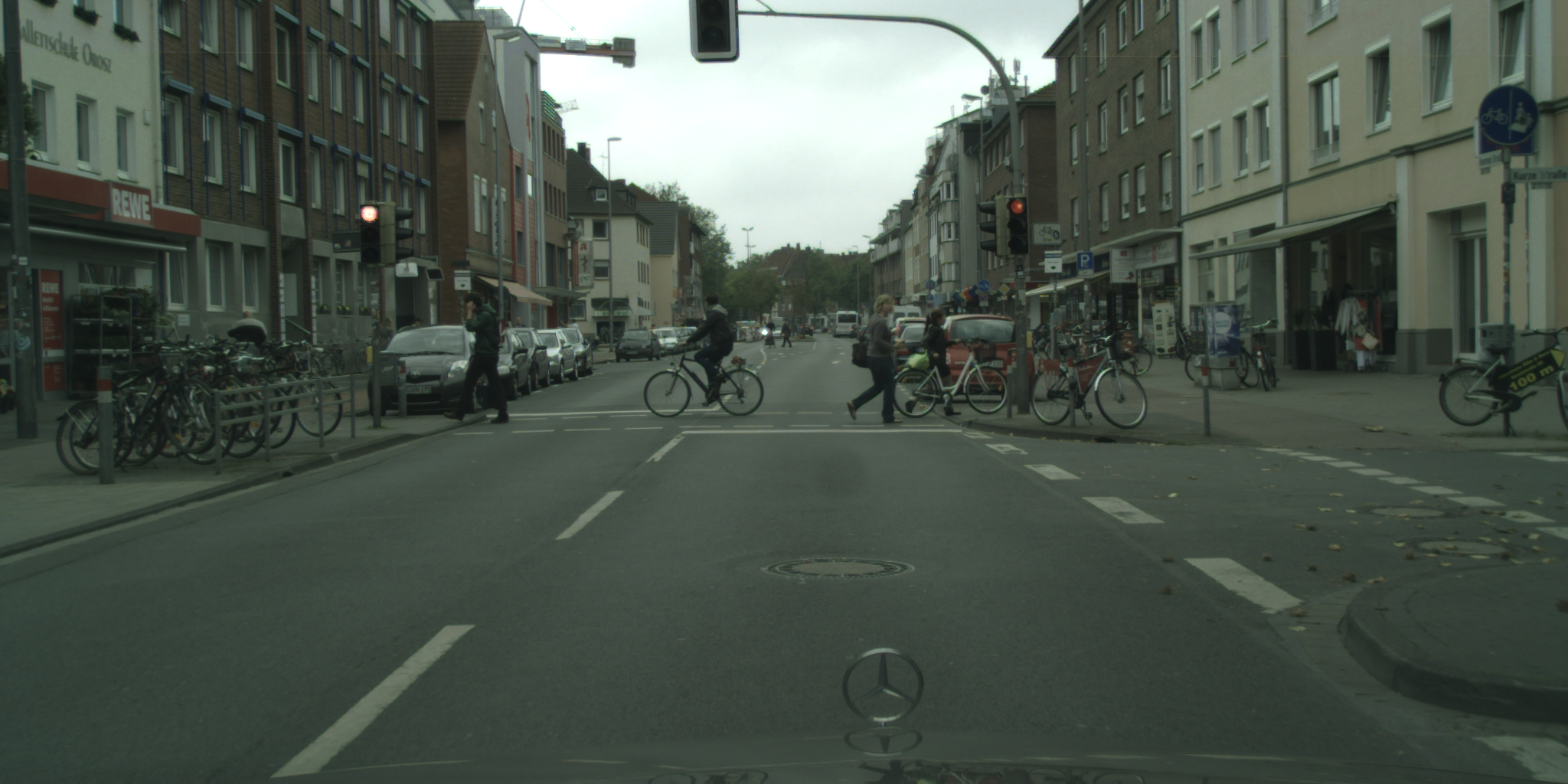}
	\end{subfigure}
	\begin{subfigure}{0.24\textwidth}
		\centering
		\includegraphics[width=4.8cm,height=2.5cm]{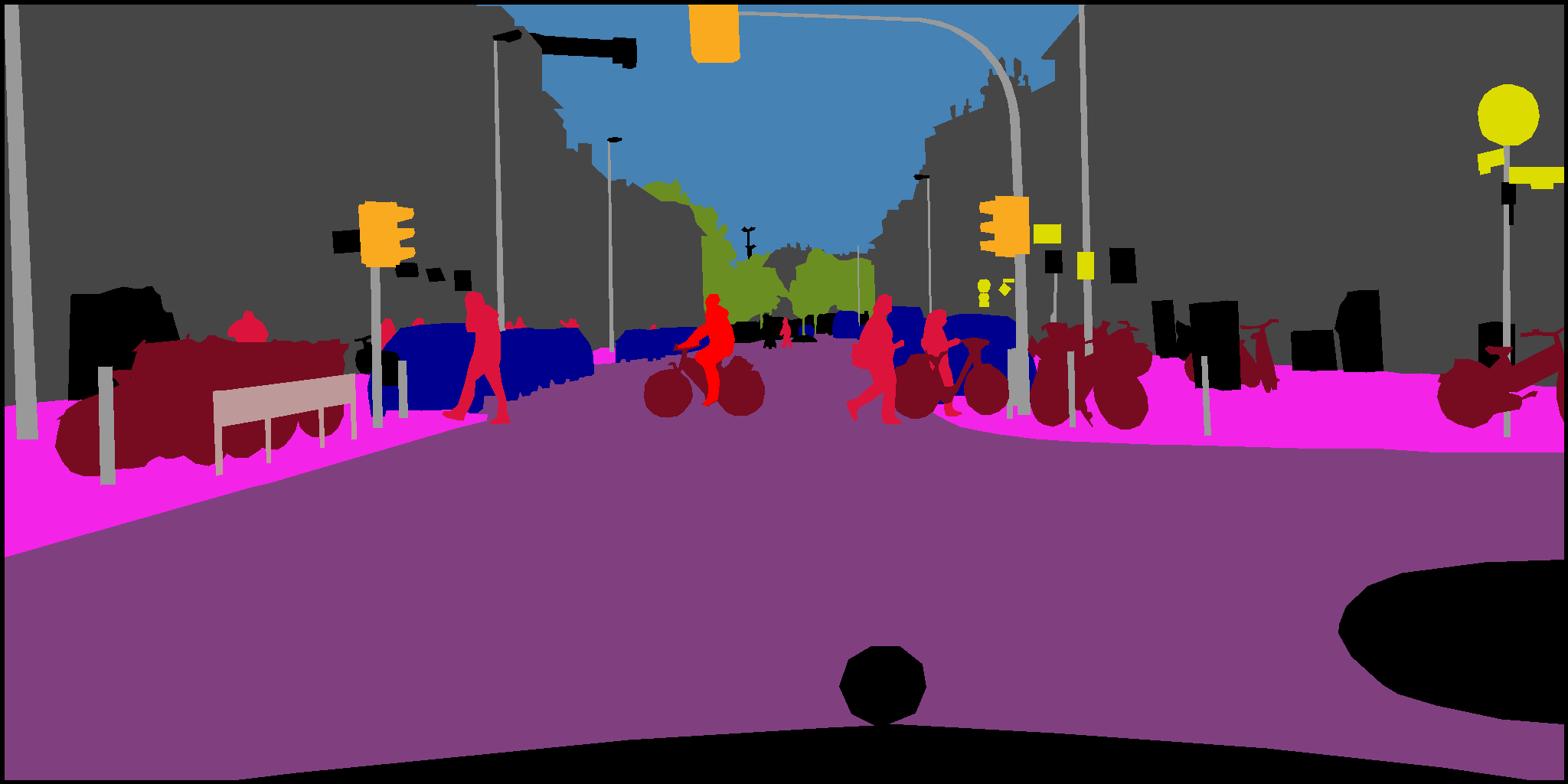}
	\end{subfigure}
	\begin{subfigure}{0.24\textwidth}
		\centering
		\includegraphics[width=4.8cm,height=2.5cm]{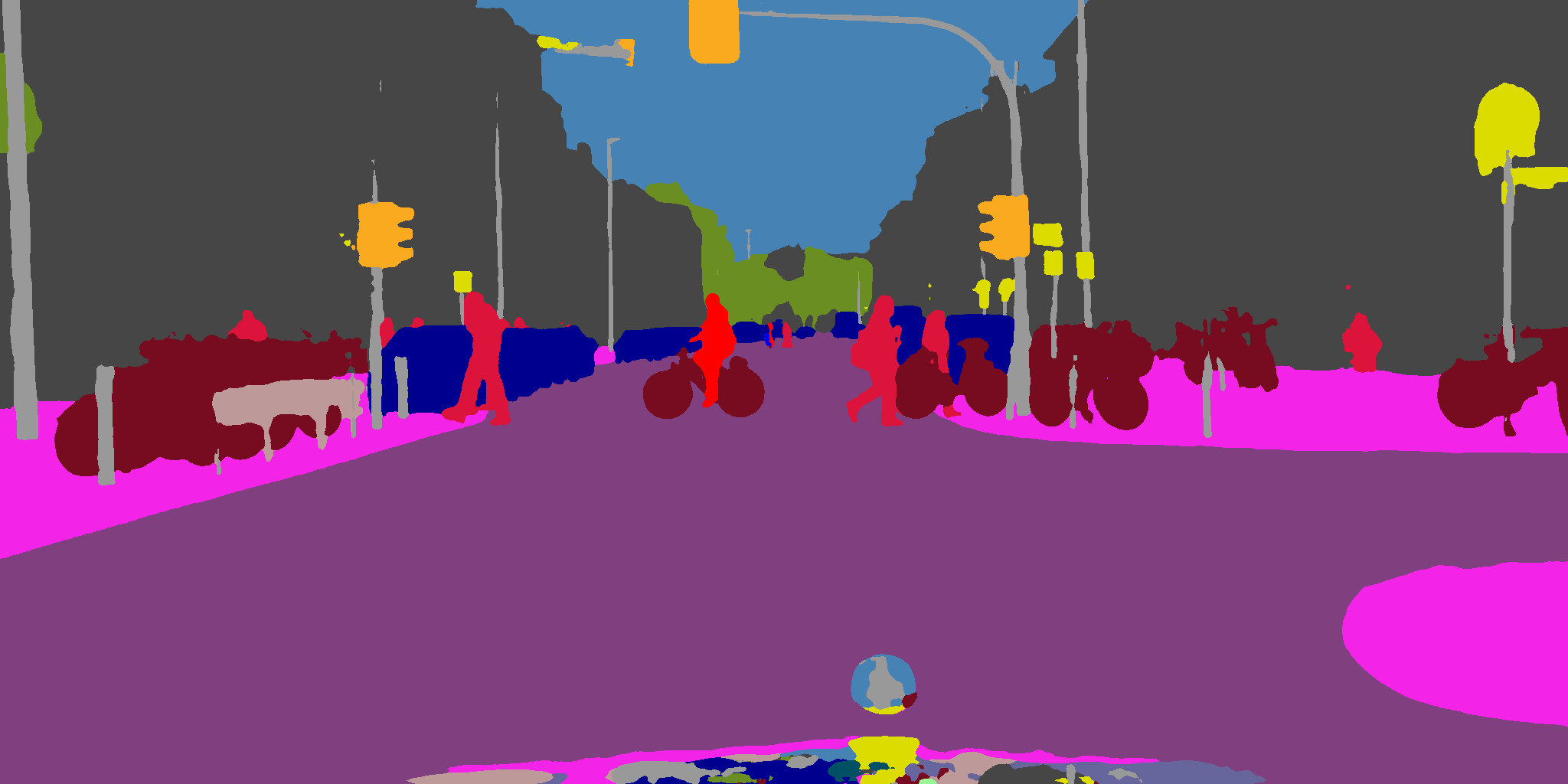}
	\end{subfigure}
	\centering
	\\
	\centering
	\begin{subfigure}{0.24\textwidth}
		\centering
		\includegraphics[width=4.8cm,height=2.5cm]{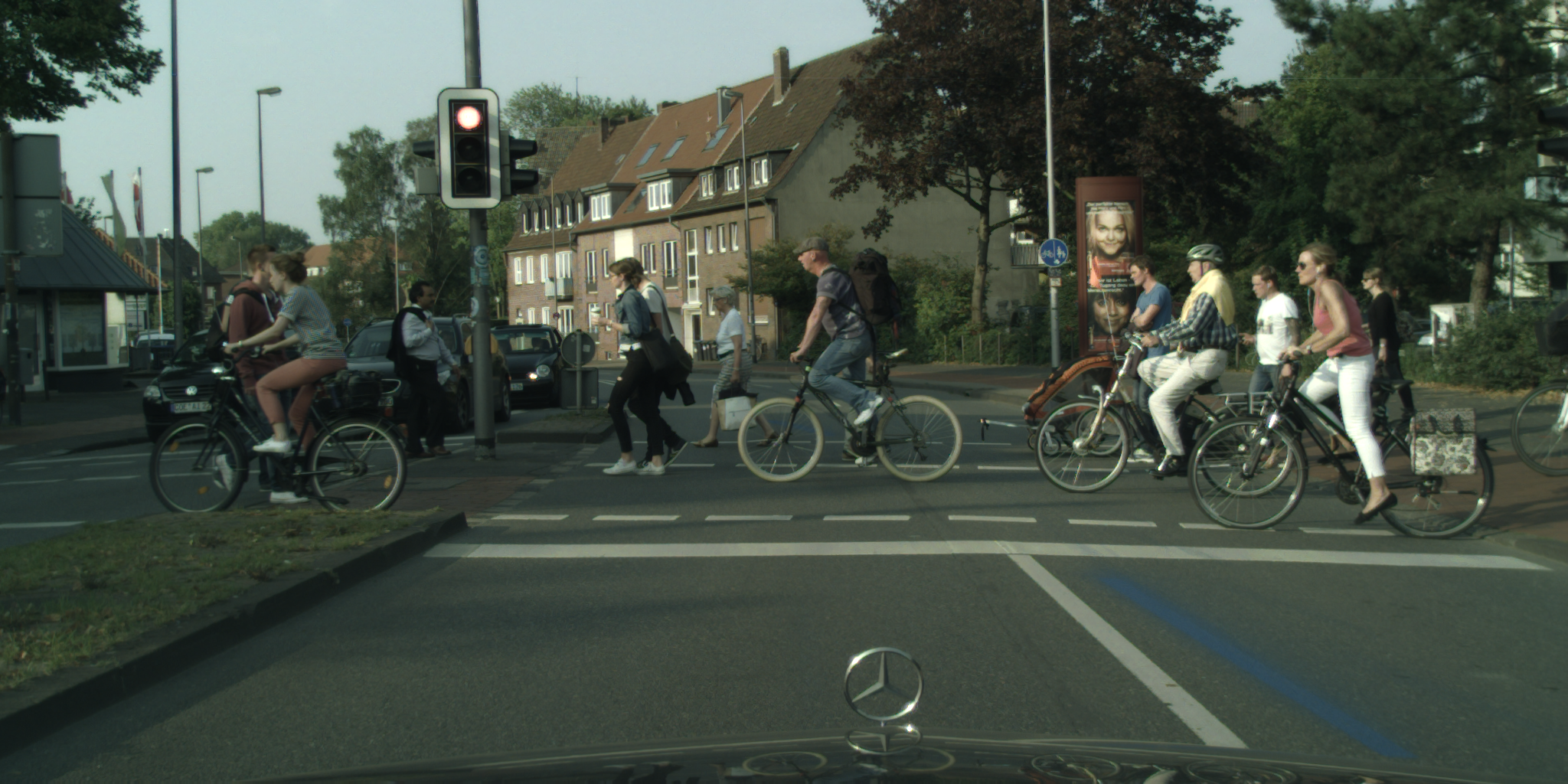}
	\end{subfigure}
	\begin{subfigure}{0.24\textwidth}
		\centering
		\includegraphics[width=4.8cm,height=2.5cm]{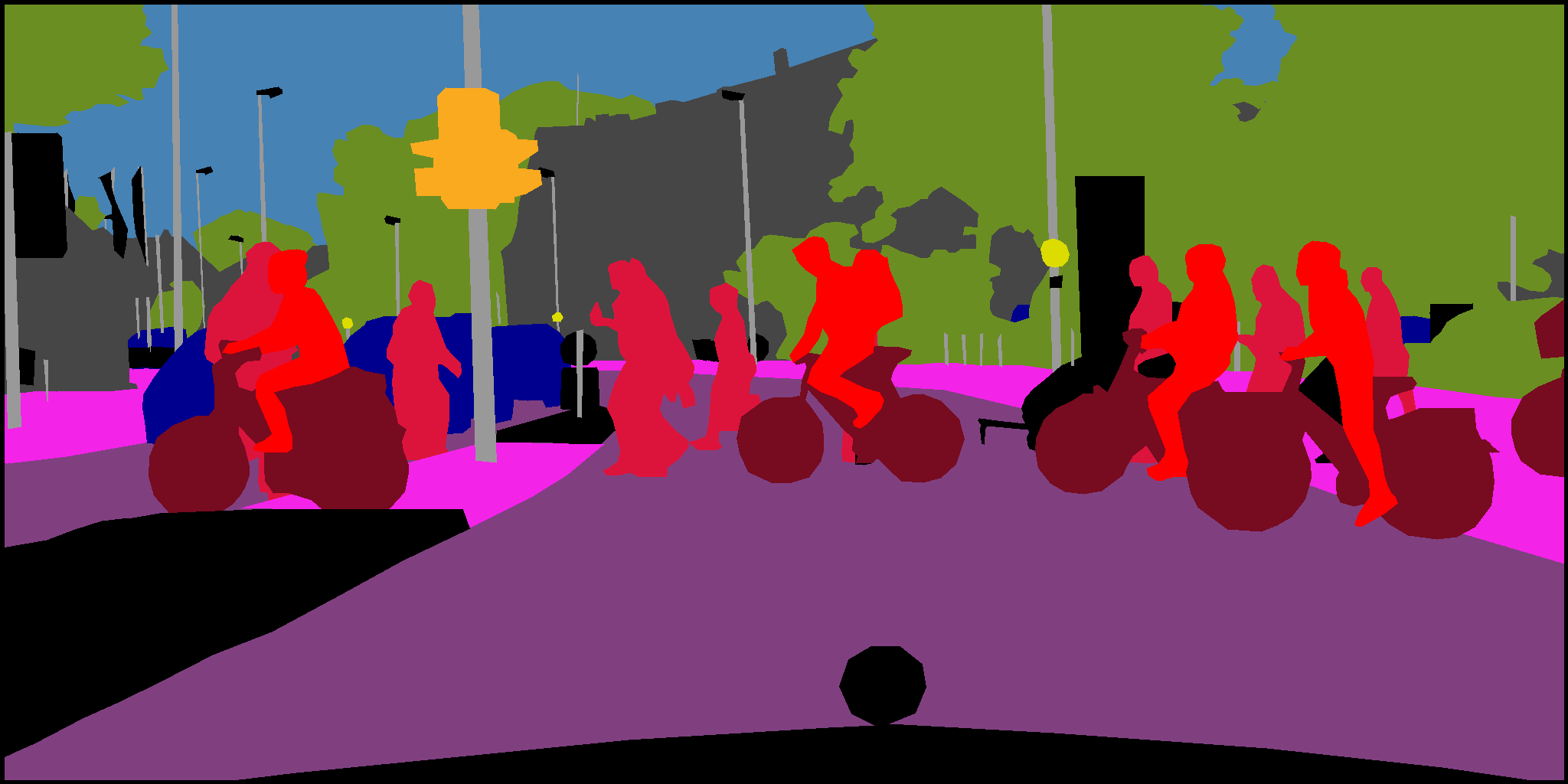}
	\end{subfigure}
	\begin{subfigure}{0.24\textwidth}
		\centering
		\includegraphics[width=4.8cm,height=2.5cm]{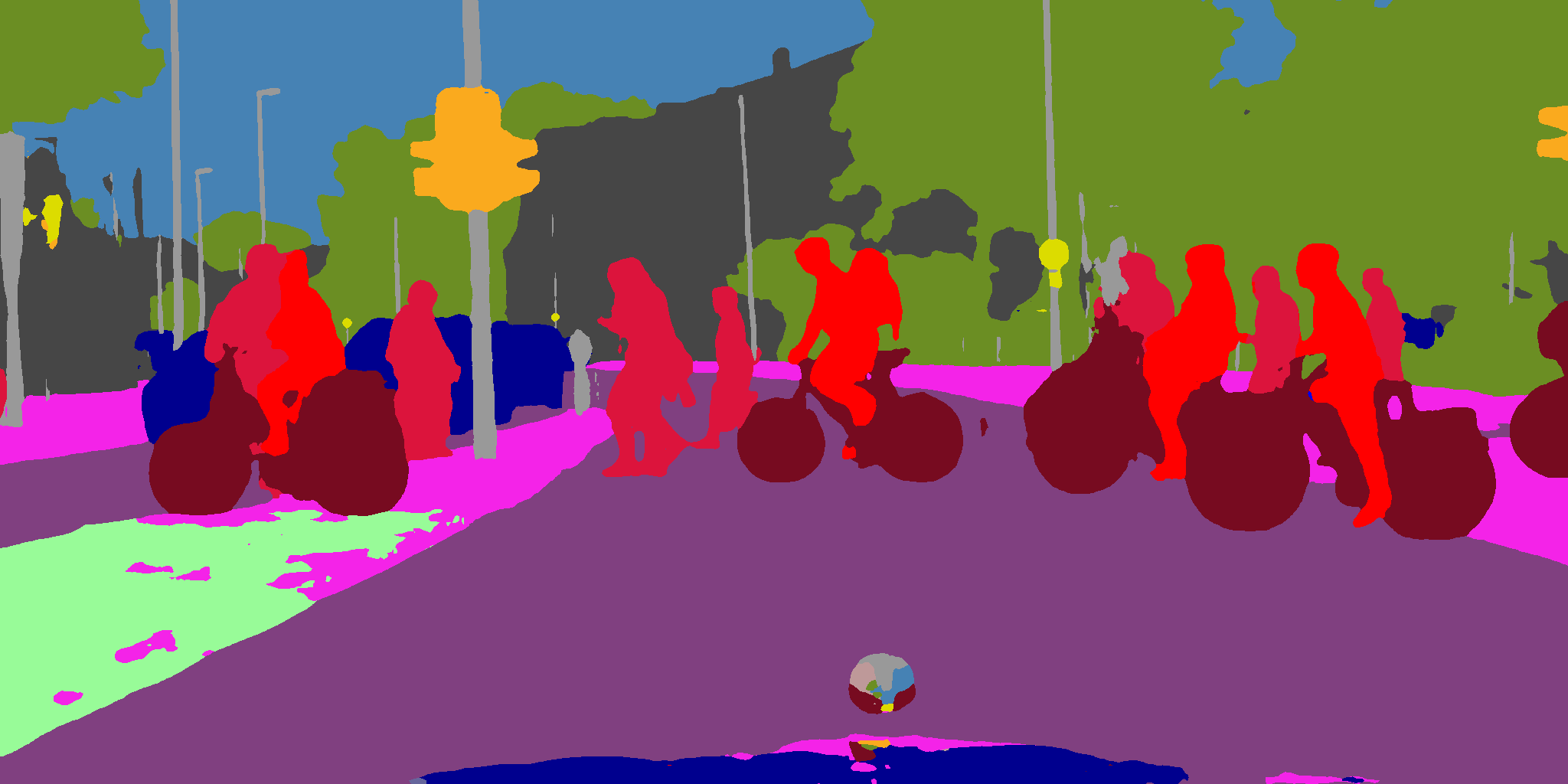}
	\end{subfigure}
	\centering
	\caption{Visual results of our method SPFNet on Cityscapes dataset. The first row is the original image,the second row is the ground truth while the last row is represents the model performance.}
	\label{cityscapes_fig}
\end{figure*}

Table \ref{tab:cityscape_result_sota} presents a comprehensive comparison of the proposed SPFNet against other state-of-the-art models on the Cityscapes dataset, considering factors such as Floating-Point Operations per Second (FLOPS), parameters, and speed. The experiments were conducted on both the training and validation sets, utilizing a reduced input resolution of 
2048
×
1024
2048×1024 to enhance training efficiency. Subsequently, model segmentation accuracy was evaluated on the test set, and the results were submitted to the Cityscapes dataset online server 
footnote{\url{https://www.cityscapes-dataset.com/submit/}} to obtain benchmark results. The comparison encompasses both smaller networks such as ICNet \cite{zhao2018icnet}, BiSeNet \cite{bilinski2018dense}, and larger networks including PSPNet \cite{zhao2017pyramid}, RefineNet \cite{lin2017refinenet}, and Deeplab \cite{chen2014semantic}. Notably, SPFNet34H achieves a mean Intersection over Union (mIoU) of 75.7\%, featuring 41.9 million parameters and an inference speed of 12.7 frames per second (FPS). On the other hand, SPFNet18L achieves a mIoU of 71.9\%, with 31.7 million parameters and an inference speed of 46.5 FPS. It is worth highlighting that SPFNet employs ResNet34 and ResNet18, which have fewer parameters compared to many larger models in Table \ref{tab:cityscape_result_sota}, yet achieve comparable results. Importantly, our models do not employ multi-scale testing or multi-crop evaluation, techniques often utilized to enhance accuracy by other practitioners. Figure \ref{cityscapes_fig} illustrates visual examples of SPFNet's performance on the Cityscapes validation set, providing qualitative insights into its segmentation capabilities.
\begin{table*}
    \centering
	\caption{\MakeUppercase{Comparison between the proposed method SPFNet and other state-of-the-arts methods on the Cityscapes test dataset.}}
	\label{tab:cityscape_result_sota}
 		\begin{adjustbox}{width=0.9\textwidth}
 			\small
			\begin{tabular}{l|c|c|c|c|c|c|c}
			\hline
	    	Method&Backbone&Resolution&Params (M)&FLOPS (G)&Speed(FPS)&test set&mIoU\\ 
			\hline
			\hline
	   	    CRF-RNN\cite{zheng2015conditional}&VGG16&512$\times$1024&-&- &1.4&\checkmark&62.5\\
			Deeplab\cite{chen2014semantic}&VGG16&512$\times$1024&262.1&457.8&0.25& \checkmark&63.1\\
			FCN\cite{long2015fully}&VGG16 &512$\times$1024&134.5&136&2&\checkmark &65.3\\
			Dilation10\cite{yu2016multi}&VGG16&512$\times$1024&140.8&-&-&\checkmark &67.1\\
			RefineNet\cite{lin2017refinenet}&ResNet101 &512$\times$1024&118.1&526&9.1&\checkmark &73.6\\
			PSPNet\cite{zhao2017pyramid}&ResNet101 &713$\times$713&250&412.2&0.78&\checkmark &78.4\\
			\hline
			SegNet\cite{badrinarayanan2017segnet}&BiseNet18&640$\times$360&29.5&286&16.7&\checkmark &57\\
			TD4-Bise18\cite{hu2020temporally}&BiseNet18&1024$\times$2048&-&&&\checkmark &74.9\\
			BiSeNet1\cite{yu2018bisenet}&Xception39&768$\times$1536&5.8&14.8&72.3&\checkmark &68.4\\
			BiSeNet2\cite{yu2018bisenet}&ResNet18&768$\times$1536&49.0&54.0&45.7&\checkmark &74.7\\
			ICNet\cite{zhao2018icnet}&PSPNet50&1024$\times$2048&26.5&28.3&30.3&\checkmark &69.5\\
			 LBN-AA\cite{dong2020real}&LBN-AA+MobileNetV2&488$\times$896&6.2&49.5&51.0&\checkmark &74.4\\
			BiSeNetV2\-L\cite{yu2021bisenet}&No&512$\times$1024&&&47.3&\checkmark &75.3\\
			\hline
			SPFNet34H&ResNet34M&512$\times$1024&41.9&317&12.7&\checkmark &75.7\\
			SPFNet18L&ResNet18&512$\times$1024&31.7&61.0&46.5&\checkmark &71.9\\
				\hline
			\end{tabular}
 		\end{adjustbox}
\end{table*}

\subsection{Results on Camvid Dataset}

Table \ref{tab:individual_camvid_s} presents the segmentation accuracy of our proposed method on the Camvid test set, offering a comparative analysis with other state-of-the-art methods. The model was trained with an input resolution of $360\times480$ pixels, utilizing only the training and validation samples for training and validation, respectively. The testing samples were reserved to obtain results for comparison with other state-of-the-art models. Notably, our model exhibits superior accuracy across various configurations. The SPFNet34H, leveraging ResNet34 as its backbone, achieves a significant 75.1\% mean Intersection over Union (mIoU) on the test set, accompanied by an inference speed of 33.0 frames per second (FPS), representing a well-balanced trade-off between speed and accuracy. The SPFNet43L, incorporating downsampling at stage 2, achieves a 71.4\% mIoU with a faster inference speed of 79.3 FPS. Additionally, we evaluate our methods with two configurations of ResNet18. SPFNet18H attains a 72.2\% mIoU with an inference speed of 39.7 FPS, while SPFNet18L achieves a 70.3\% mIoU with a higher inference speed of 109 FPS. These experiments validate the efficacy of the proposed architecture design. As demonstrated in Table \ref{tab:individual_camvid_s}, training the model with extra data enhances performance for some methods. For instance, BiSeNetV2* \cite{yu2021bisenet}, when trained without extra data, achieves a 72.4\% mIoU, and its performance improves to 76.7\% when the model is initially trained on Cityscapes data. Similarly, BiSeNetV2-L* improves from 73.2\% mIoU, without extra data, to 78.5\% with the same settings. Individual category results on the Camvid test set are reported in Table \ref{tab:individual_camvid_s}. Finally, we showcase visual results of our SPFNet method on the Camvid test set.

\begin{table}
	\centering
	\caption{\MakeUppercase{The accuracy, speed and parameters comparison of the proposed method against other semantic segmentation methods on the Camvid test set. $^*$ indicates the model pre-trained on Cityscapes. ”-” indicates the corresponding result is not reported by the methods.}}
	\label{tab:sota_camvid_s}
	\vspace{1ex}
	\begin{adjustbox}{width=0.48\textwidth}
		\small
		\begin{tabular}{l|c|c|c|c|c}
			\hline 
			Method&Year&Resolution&Params (M)&Speed (FPS)&mIoU\\
			\hline\hline
			Deeplabv2\cite{chen2017deeplab}&2017&$720\times960$&262.1&4.9&61.6\\
			PSPNet\cite{zhao2017pyramid}&2017&$720\times960$&250&5.4&69.1\\
			DenseDecoder\cite{bilinski2018dense}&2018&$720\times960$&-&-&70.9\\
			Dilation8\cite{yu2015multi}&2016&$720\times960$&-&4.4&65.3\\
			\hline
			SegNet\cite{badrinarayanan2017segnet}&2015&$360\times480$&29.5&-&55.6\\
			ENet\cite{paszke2016enet}&2016&$360\times480$&0.37&-&51.3\\
			DFANet-A\cite{li2019dfanet}&2019&$720\times960$&7.8&120&64.7\\
			DFANet-B\cite{li2019dfanet}&2019&$720\times960$&4.8&160&59.3\\
			BiSeNet1\cite{yu2018bisenet}&2018&$720\times960$&5.8&175&65.7\\
			BiSeNet2\cite{yu2018bisenet}&2018&$720\times960$&49.0&116.3&68.7\\
			ICNet\cite{zhao2018icnet}&2018&$720\times960$&26.5&27.8&67.1\\
			DABNet\cite{li2019dabnet}&2019&$360\times480$&0.76&&66.4\\
			CAS\cite{zhang2019customizable}&2020&$720\times960$&-&169&71.2\\
			GAS\cite{lin2020graph}&2020&$720\times960$&-&153.1&72.8\\
			AGLNet\cite{zhou2020aglnet}&2020&$360\times480$&1.12&90.1&69.4\\
			CGNet\cite{9292449cgnet}&2020&$360\times480$&0.5&-&65.6\\
			NDNet45-FCN8-LF\cite{yang2020small}&2020&$360\times480$&1.1&-&57.5\\
			LBN-AA\cite{dong2020real}&2020&$720\times960$&6.2&39.3&68.0\\
			BiSeNetV2/BiSeNetV2L\cite{yu2021bisenet}&2021&$720\times960$&-&124.5/32.7&72.4/73.2\\
			BiSeNetV2$^*$/BiSeNetV2L$^*$\cite{yu2021bisenet}&2021&$720\times960$&-&124.5/32.7&76.7/78.5\\
			\hline
			SPFNet34H(ours) & &$360\times480$&41.8&33&75.1\\
			SPFNet34L(ours) & &$360\times480$&41.8&79.3&71.4\\
			SPFNet18H(ours) & &$360\times480$&31.7 &39.7&72.2\\
			SPFNet18L(ours) & &$360\times480$&31.7&109&70.3\\
			\hline
		\end{tabular}
	\end{adjustbox}
\end{table}

\begin{table*}
	\centering
	\caption{\MakeUppercase{Individual category results on Camvid test set in terms of mIoU for 11 classes.
	”-” indicates the corresponding result is not reported by the methods.}}
	\label{tab:individual_camvid_s}
	\vspace{1ex}
	\begin{tabu}{l|c|c|c|c|c|c|c|c|c|c|c|c}
		\hline
		Method&Building&Tree&Sky&Car&Sign&Road&Ped&Fence&Pole&Sidewalk&Bicyclist&mIoU\\
		\hline
	\hline
	SegNet\cite{badrinarayanan2017segnet}&\textbf{88.8}&\textbf{87.3}&92.4&82.1&20.5&97.2&57.1&49.3&27.5&84.4&30.7&65.2\\
	BiSeNet1\cite{yu2018bisenet}&82.2&74.4&91.9&80.8&42.8&93.3&53.8&49.7&25.4&77.3&50.0&65.6\\
	BiSeNet2\cite{yu2018bisenet}&83.0&75.8&92.0&83.7&46.5&94.6&58.8&53.6&31.9&81.4&54.0&68.7\\
	AGLNet\cite{zhou2020aglnet}&82.6&76.1&91.8&87.0&45.3&95.4&61.5&39.5&39.0&83.1&62.7&69.4\\
	LBN-AA\cite{dong2020real}&83.2&70.5&92.5&81.7&51.6&93.0&55.6&53.2&36.3&82.1&47.9&68.0\\
	BiSeNetV2\slash BiSeNetV2L\cite{yu2021bisenet}&-&-&-&-&-&-&-&-&-&-&-&72.4\slash 73.2\\ \hline
	SPFNet34H(ours)&86.4&79.2&\textbf{92.6}&\textbf{90.3}&\textbf{56.3}&\textbf{95.9}&\textbf{68.1}&\textbf{54.5}&\textbf{42.3}&85.4&\textbf{74.2}&\textbf{75.1}\\ 
	SPFNet34L(ours)&84.6&77.2&91.6&90.2&49.9&95.6&61.5&47.9&35.4&85.0&66.1&71.4\\ 
	SPFNet18H(ours)&84.7&78.0&92.5&89.9&52.0&95.4&66.0&43.1&40.7&84.1&67.7&72.2\\ 
	SPFNet18L(ours)&83.7&76.7&91.8&89.6&49&95.4&59.7&45.2&34.4&83.5&63.7&70.3\\
	\hline
	\end{tabu}
\end{table*}  
\begin{figure*}
	\centering
	\begin{subfigure}{0.24\textwidth}
		\centering
		\includegraphics[width=4.8cm,height=2cm]{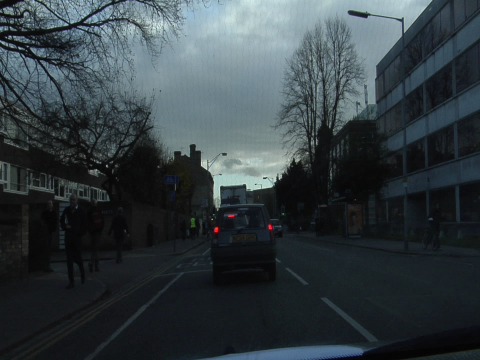}
	\end{subfigure}
	\begin{subfigure}{0.24\textwidth}
		\centering
		\includegraphics[width=4.8cm,height=2cm]{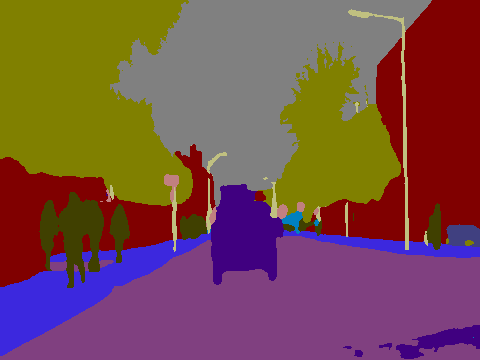}
	\end{subfigure}
	\begin{subfigure}{0.24\textwidth}
		\centering
		\includegraphics[width=4.8cm,height=2cm]{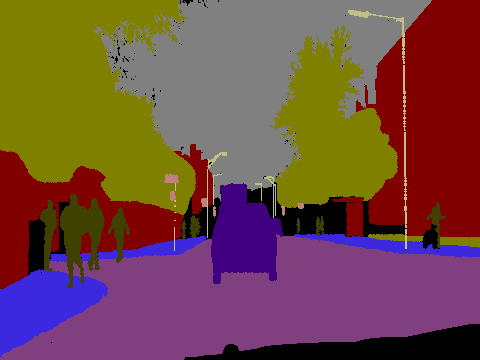}
	\end{subfigure}
	\\
	\begin{subfigure}{0.24\textwidth}
		\centering
		\includegraphics[width=4.8cm,height=2cm]{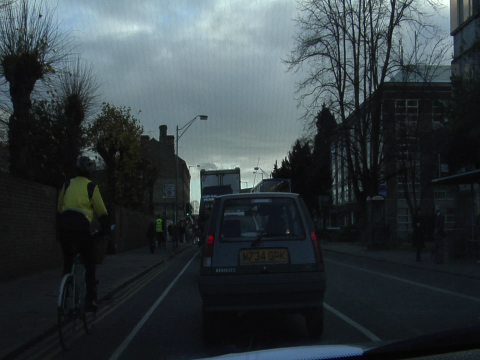}
	\end{subfigure}
	\begin{subfigure}{0.24\textwidth}
		\centering
		\includegraphics[width=4.8cm,height=2cm]{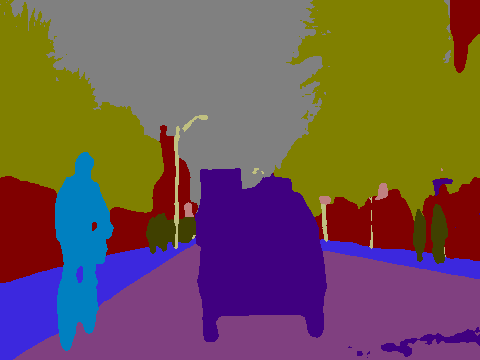}
	\end{subfigure}
	\begin{subfigure}{0.24\textwidth}
		\centering
		\includegraphics[width=4.8cm,height=2cm]{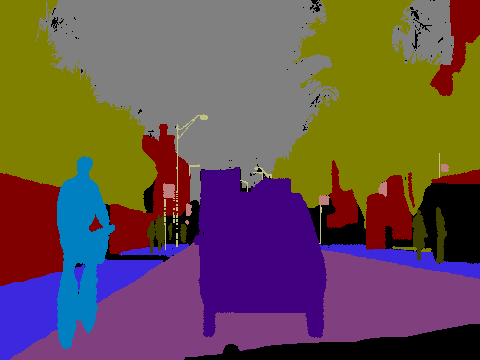}
	\end{subfigure}
	\\
	\centering
	\begin{subfigure}{0.24\textwidth}
		\centering
		\includegraphics[width=4.8cm,height=2cm]{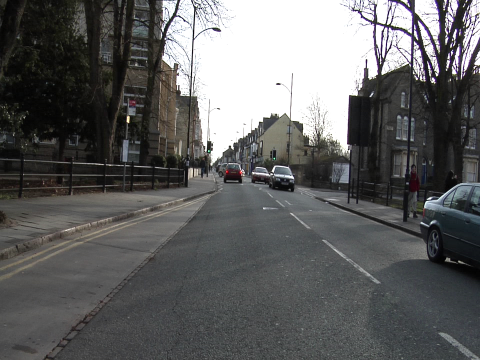}
	\end{subfigure}
	\begin{subfigure}{0.24\textwidth}
		\centering
		\includegraphics[width=4.8cm,height=2cm]{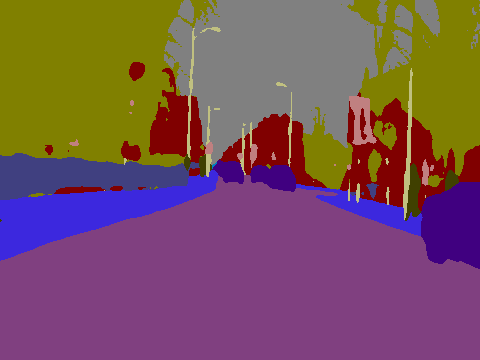}
	\end{subfigure}
	\begin{subfigure}{0.24\textwidth}
		\centering
		\includegraphics[width=4.8cm,height=2cm]{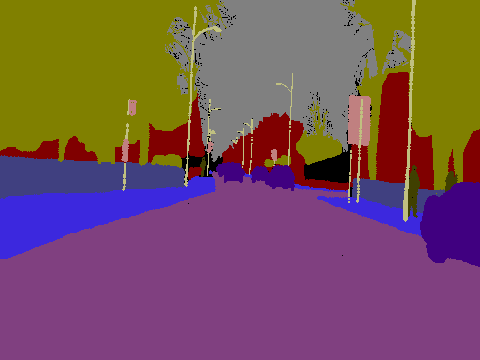}
	\end{subfigure}
	\\
	\centering
	\begin{subfigure}{0.24\textwidth}
		\centering
		\includegraphics[width=4.8cm,height=2cm]{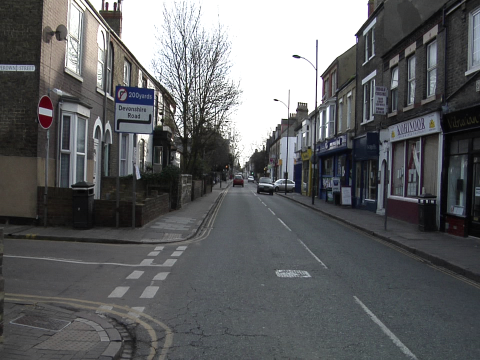}
	\end{subfigure}
	\begin{subfigure}{0.24\textwidth}
		\centering
		\includegraphics[width=4.8cm,height=2cm]{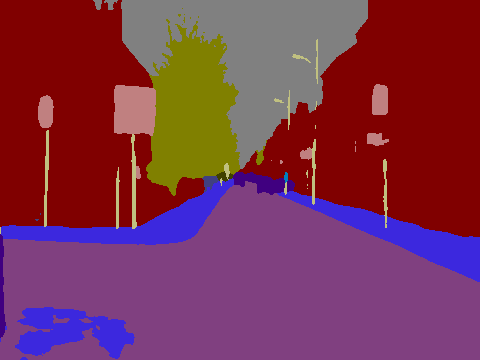}
	\end{subfigure}
	\begin{subfigure}{0.24\textwidth}
		\centering
		\includegraphics[width=4.8cm,height=2cm]{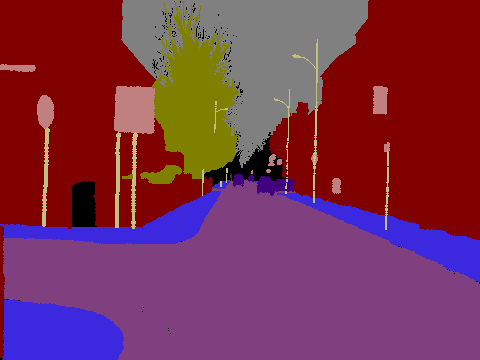}
	\end{subfigure}
	\\
	\centering
	\begin{subfigure}{0.24\textwidth}
		\centering
		\includegraphics[width=4.8cm,height=2cm]{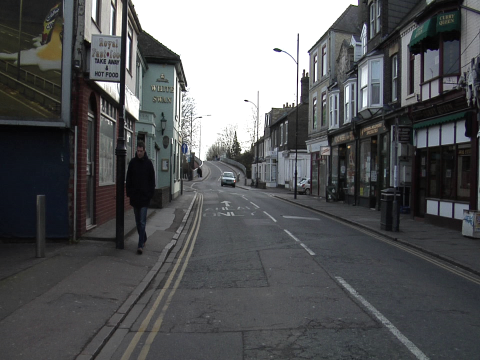}
	\end{subfigure}
	\begin{subfigure}{0.24\textwidth}
		\centering
		\includegraphics[width=4.8cm,height=2cm]{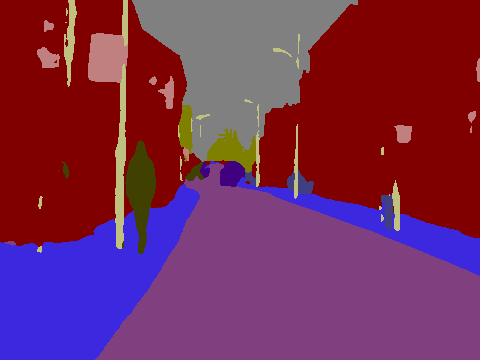}
	\end{subfigure}
	\begin{subfigure}{0.24\textwidth}
		\centering
		\includegraphics[width=4.8cm,height=2cm]{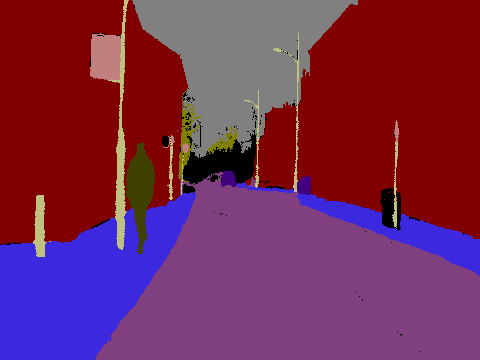}
	\end{subfigure}
	\\
	\centering
	\begin{subfigure}{0.24\textwidth}
		\centering
		\includegraphics[width=4.8cm,height=2cm]{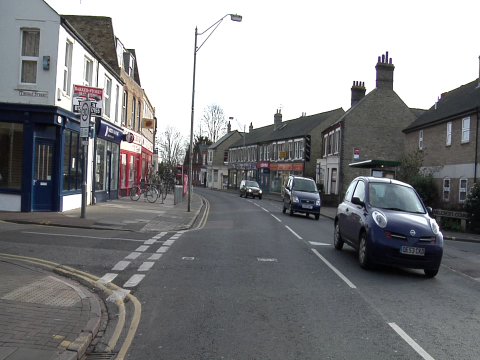}
	\end{subfigure}
	\begin{subfigure}{0.24\textwidth}
		\centering
		\includegraphics[width=4.8cm,height=2cm]{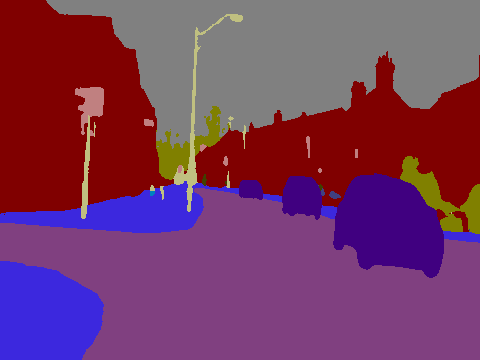}
	\end{subfigure}
	\begin{subfigure}{0.24\textwidth}
		\centering
		\includegraphics[width=4.8cm,height=2cm]{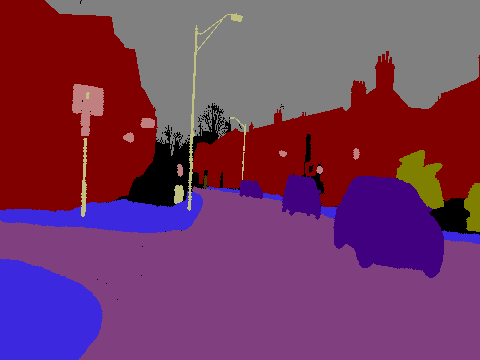}
	\end{subfigure}
	\caption{Visual results of our method SPFNet on Camvid test set. The first row is the image,the second row is the prediction while the last row is ground truth.}
	\label{Figure:Fig6}
\end{figure*}

\section{Conclusion}
This paper introduces the Subspace Pyramid Fusion Module (SPFM), a novel method designed for learning multi-scale context information by partitioning input feature maps into distinct subspaces. Additionally, we propose the Efficient Shuffle Attention Module (ESAM), which leverages the channel-shuffle operator to facilitate enhanced information communication across various sub-features. Combining these innovations, we present a semantic segmentation architecture, termed SPFNet, specifically tailored to address the challenges of multi-scale feature fusion. Ablation studies conducted on the Cityscapes dataset underscore the effectiveness of the proposed SPFM and ESAM modules. The performance evaluation on the Camvid dataset demonstrates the efficacy of SPFNet across various configurations. Specifically, SPFNet34H achieves a mean Intersection over Union (mIoU) of 75.1\%, SPFNet34L achieves 71.4\% mIoU, SPFNet18H achieves 72.2\% mIoU, and SPFNet18L achieves 70.3\% mIoU. Furthermore, SPFNet34H and SPFNet18L exhibit competitive results on the Cityscapes test sets, achieving mIoUs of 75.7\% and 71.9\%, respectively. These findings underscore the potential of SPFNet in addressing semantic segmentation tasks with improved accuracy and efficiency.
\ifCLASSOPTIONcaptionsoff
\clearpage
\fi
\bibliographystyle{IEEEtran}
\bibliography{refs}

\begin{thebibliography}{10}
\providecommand{\url}[1]{#1}
\csname url@samestyle\endcsname
\providecommand{\newblock}{\relax}
\providecommand{\bibinfo}[2]{#2}
\providecommand{\BIBentrySTDinterwordspacing}{\spaceskip=0pt\relax}
\providecommand{\BIBentryALTinterwordstretchfactor}{4}
\providecommand{\BIBentryALTinterwordspacing}{\spaceskip=\fontdimen2\font plus
\BIBentryALTinterwordstretchfactor\fontdimen3\font minus
  \fontdimen4\font\relax}
\providecommand{\BIBforeignlanguage}[2]{{%
\expandafter\ifx\csname l@#1\endcsname\relax
\typeout{** WARNING: IEEEtran.bst: No hyphenation pattern has been}%
\typeout{** loaded for the language `#1'. Using the pattern for}%
\typeout{** the default language instead.}%
\else
\language=\csname l@#1\endcsname
\fi
#2}}
\providecommand{\BIBdecl}{\relax}
\BIBdecl

\bibitem{ronneberger2015u}
O.~Ronneberger, P.~Fischer, and T.~Brox, ``U-net: Convolutional networks for
  biomedical image segmentation,'' in \emph{International Conference on Medical
  image computing and computer-assisted intervention}.\hskip 1em plus 0.5em
  minus 0.4em\relax Springer, 2015, pp. 234--241.

\bibitem{saha2018her2net}
M.~Saha and C.~Chakraborty, ``Her2net: A deep framework for semantic
  segmentation and classification of cell membranes and nuclei in breast cancer
  evaluation,'' \emph{IEEE Transactions on Image Processing}, vol.~27, no.~5,
  pp. 2189--2200, 2018.

\bibitem{siam2018comparative}
M.~Siam, M.~Gamal, M.~Abdel-Razek, S.~Yogamani, M.~Jagersand, and H.~Zhang, ``A
  comparative study of real-time semantic segmentation for autonomous
  driving,'' in \emph{Proceedings of the IEEE conference on computer vision and
  pattern recognition workshops}, 2018, pp. 587--597.

\bibitem{zhao2018psanet}
H.~Zhao, Y.~Zhang, S.~Liu, J.~Shi, C.~C. Loy, D.~Lin, and J.~Jia, ``Psanet:
  Point-wise spatial attention network for scene parsing,'' in
  \emph{Proceedings of the European Conference on Computer Vision (ECCV)},
  2018, pp. 267--283.

\bibitem{zhao2017pyramid}
H.~Zhao, J.~Shi, X.~Qi, X.~Wang, and J.~Jia, ``Pyramid scene parsing network,''
  in \emph{Proceedings of the IEEE conference on computer vision and pattern
  recognition}, 2017, pp. 2881--2890.

\bibitem{long2015fully}
J.~Long, E.~Shelhamer, and T.~Darrell, ``Fully convolutional networks for
  semantic segmentation,'' in \emph{Proceedings of the IEEE conference on
  computer vision and pattern recognition}, 2015, pp. 3431--3440.

\bibitem{badrinarayanan2017segnet}
V.~Badrinarayanan, A.~Kendall, and R.~Cipolla, ``Segnet: A deep convolutional
  encoder-decoder architecture for image segmentation,'' \emph{IEEE
  transactions on pattern analysis and machine intelligence}, vol.~39, no.~12,
  pp. 2481--2495, 2017.

\bibitem{zhou2018unet++}
Z.~Zhou, M.~M.~R. Siddiquee, N.~Tajbakhsh, and J.~Liang, ``Unet++: A nested
  u-net architecture for medical image segmentation,'' in \emph{Deep learning
  in medical image analysis and multimodal learning for clinical decision
  support}.\hskip 1em plus 0.5em minus 0.4em\relax Springer, 2018, pp. 3--11.

\bibitem{yu2015multi}
F.~Yu and V.~Koltun, ``Multi-scale context aggregation by dilated
  convolutions,'' \emph{arXiv preprint arXiv:1511.07122}, 2015.

\bibitem{chen2017deeplab}
L.-C. Chen, G.~Papandreou, I.~Kokkinos, K.~Murphy, and A.~L. Yuille, ``Deeplab:
  Semantic image segmentation with deep convolutional nets, atrous convolution,
  and fully connected crfs,'' \emph{IEEE transactions on pattern analysis and
  machine intelligence}, vol.~40, no.~4, pp. 834--848, 2017.

\bibitem{sun2019deep}
K.~Sun, B.~Xiao, D.~Liu, and J.~Wang, ``Deep high-resolution representation
  learning for human pose estimation,'' in \emph{Proceedings of the IEEE/CVF
  Conference on Computer Vision and Pattern Recognition}, 2019, pp. 5693--5703.

\bibitem{chen2018encoder}
L.-C. Chen, Y.~Zhu, G.~Papandreou, F.~Schroff, and H.~Adam, ``Encoder-decoder
  with atrous separable convolution for semantic image segmentation,'' in
  \emph{Proceedings of the European conference on computer vision (ECCV)},
  2018, pp. 801--818.

\bibitem{li2018pyramid}
H.~Li, P.~Xiong, J.~An, and L.~Wang, ``Pyramid attention network for semantic
  segmentation,'' \emph{arXiv preprint arXiv:1805.10180}, 2018.

\bibitem{yang2018denseaspp}
M.~Yang, K.~Yu, C.~Zhang, Z.~Li, and K.~Yang, ``Denseaspp for semantic
  segmentation in street scenes,'' in \emph{Proceedings of the IEEE conference
  on computer vision and pattern recognition}, 2018, pp. 3684--3692.

\bibitem{lin2017refinenet}
G.~Lin, A.~Milan, C.~Shen, and I.~Reid, ``Refinenet: Multi-path refinement
  networks for high-resolution semantic segmentation,'' in \emph{Proceedings of
  the IEEE conference on computer vision and pattern recognition}, 2017, pp.
  1925--1934.

\bibitem{elhassan2021ppanet}
M.~A. Elhassan, Y.~Chen, Y.~Chen, C.~Huang, J.~Yang, X.~Yao, C.~Yang, and
  Y.~Cheng, ``Ppanet: Point-wise pyramid attention network for semantic
  segmentation,'' \emph{Wireless Communications and Mobile Computing}, vol.
  2021, 2021.

\bibitem{liu2015parsenet}
W.~Liu, A.~Rabinovich, and A.~C. Berg, ``Parsenet: Looking wider to see
  better,'' \emph{arXiv preprint arXiv:1506.04579}, 2015.

\bibitem{elhassan2021dsanet}
M.~A. Elhassan, C.~Huang, C.~Yang, and T.~L. Munea, ``Dsanet: Dilated spatial
  attention for real-time semantic segmentation in urban street scenes,''
  \emph{Expert Systems with Applications}, vol. 183, p. 115090, 2021.

\bibitem{yu2018bisenet}
C.~Yu, J.~Wang, C.~Peng, C.~Gao, G.~Yu, and N.~Sang, ``Bisenet: Bilateral
  segmentation network for real-time semantic segmentation,'' in
  \emph{Proceedings of the European conference on computer vision (ECCV)},
  2018, pp. 325--341.

\bibitem{iandola2016squeezenet}
F.~N. Iandola, S.~Han, M.~W. Moskewicz, K.~Ashraf, W.~J. Dally, and K.~Keutzer,
  ``Squeezenet: Alexnet-level accuracy with 50x fewer parameters and< 0.5 mb
  model size,'' \emph{arXiv preprint arXiv:1602.07360}, 2016.

\bibitem{howard2017mobilenets}
A.~G. Howard, M.~Zhu, B.~Chen, D.~Kalenichenko, W.~Wang, T.~Weyand,
  M.~Andreetto, and H.~Adam, ``Mobilenets: Efficient convolutional neural
  networks for mobile vision applications,'' \emph{arXiv preprint
  arXiv:1704.04861}, 2017.

\bibitem{ma2018shufflenet}
N.~Ma, X.~Zhang, H.-T. Zheng, and J.~Sun, ``Shufflenet v2: Practical guidelines
  for efficient cnn architecture design,'' in \emph{Proceedings of the European
  conference on computer vision (ECCV)}, 2018, pp. 116--131.

\bibitem{chollet2017xception}
F.~Chollet, ``Xception: Deep learning with depthwise separable convolutions,''
  in \emph{Proceedings of the IEEE conference on computer vision and pattern
  recognition}, 2017, pp. 1251--1258.

\bibitem{fan2021rethinking}
M.~Fan, S.~Lai, J.~Huang, X.~Wei, Z.~Chai, J.~Luo, and X.~Wei, ``Rethinking
  bisenet for real-time semantic segmentation,'' in \emph{Proceedings of the
  IEEE/CVF Conference on Computer Vision and Pattern Recognition}, 2021, pp.
  9716--9725.

\bibitem{li2019dabnet}
G.~Li, I.~Yun, J.~Kim, and J.~Kim, ``Dabnet: Depth-wise asymmetric bottleneck
  for real-time semantic segmentation,'' \emph{arXiv preprint
  arXiv:1907.11357}, 2019.

\bibitem{romera2017erfnet}
E.~Romera, J.~M. Alvarez, L.~M. Bergasa, and R.~Arroyo, ``Erfnet: Efficient
  residual factorized convnet for real-time semantic segmentation,'' \emph{IEEE
  Transactions on Intelligent Transportation Systems}, vol.~19, no.~1, pp.
  263--272, 2017.

\bibitem{li2019dfanet}
H.~Li, P.~Xiong, H.~Fan, and J.~Sun, ``Dfanet: Deep feature aggregation for
  real-time semantic segmentation,'' in \emph{Proceedings of the IEEE/CVF
  Conference on Computer Vision and Pattern Recognition}, 2019, pp. 9522--9531.

\bibitem{yu2021bisenet}
C.~Yu, C.~Gao, J.~Wang, G.~Yu, C.~Shen, and N.~Sang, ``Bisenet v2: Bilateral
  network with guided aggregation for real-time semantic segmentation,''
  \emph{International Journal of Computer Vision}, pp. 1--18, 2021.

\bibitem{brostow2009semantic}
G.~J. Brostow, J.~Fauqueur, and R.~Cipolla, ``Semantic object classes in video:
  A high-definition ground truth database,'' \emph{Pattern Recognition
  Letters}, vol.~30, no.~2, pp. 88--97, 2009.

\bibitem{cordts2016cityscapes}
M.~Cordts, M.~Omran, S.~Ramos, T.~Rehfeld, M.~Enzweiler, R.~Benenson,
  U.~Franke, S.~Roth, and B.~Schiele, ``The cityscapes dataset for semantic
  urban scene understanding,'' in \emph{Proceedings of the IEEE conference on
  computer vision and pattern recognition}, 2016, pp. 3213--3223.

\bibitem{ghiasi2019fpn}
G.~Ghiasi, T.-Y. Lin, and Q.~V. Le, ``Nas-fpn: Learning scalable feature
  pyramid architecture for object detection,'' in \emph{Proceedings of the
  IEEE/CVF Conference on Computer Vision and Pattern Recognition}, 2019, pp.
  7036--7045.

\bibitem{liu2018path}
S.~Liu, L.~Qi, H.~Qin, J.~Shi, and J.~Jia, ``Path aggregation network for
  instance segmentation,'' in \emph{Proceedings of the IEEE conference on
  computer vision and pattern recognition}, 2018, pp. 8759--8768.

\bibitem{tan2020efficientdet}
M.~Tan, R.~Pang, and Q.~V. Le, ``Efficientdet: Scalable and efficient object
  detection,'' in \emph{Proceedings of the IEEE/CVF conference on computer
  vision and pattern recognition}, 2020, pp. 10\,781--10\,790.

\bibitem{kirillov2019panoptic}
A.~Kirillov, R.~Girshick, K.~He, and P.~Doll{\'a}r, ``Panoptic feature pyramid
  networks,'' in \emph{Proceedings of the IEEE/CVF Conference on Computer
  Vision and Pattern Recognition}, 2019, pp. 6399--6408.

\bibitem{lin2017feature}
T.-Y. Lin, P.~Doll{\'a}r, R.~Girshick, K.~He, B.~Hariharan, and S.~Belongie,
  ``Feature pyramid networks for object detection,'' in \emph{Proceedings of
  the IEEE conference on computer vision and pattern recognition}, 2017, pp.
  2117--2125.

\bibitem{sandler2018mobilenetv2}
M.~Sandler, A.~Howard, M.~Zhu, A.~Zhmoginov, and L.-C. Chen, ``Mobilenetv2:
  Inverted residuals and linear bottlenecks,'' in \emph{Proceedings of the IEEE
  conference on computer vision and pattern recognition}, 2018, pp. 4510--4520.

\bibitem{zhang2018shufflenet}
X.~Zhang, X.~Zhou, M.~Lin, and J.~Sun, ``Shufflenet: An extremely efficient
  convolutional neural network for mobile devices,'' in \emph{Proceedings of
  the IEEE conference on computer vision and pattern recognition}, 2018, pp.
  6848--6856.

\bibitem{he2016deep}
K.~He, X.~Zhang, S.~Ren, and J.~Sun, ``Deep residual learning for image
  recognition,'' in \emph{Proceedings of the IEEE conference on computer vision
  and pattern recognition}, 2016, pp. 770--778.

\bibitem{paszke2016enet}
A.~Paszke, A.~Chaurasia, S.~Kim, and E.~Culurciello, ``Enet: A deep neural
  network architecture for real-time semantic segmentation,'' \emph{arXiv
  preprint arXiv:1606.02147}, 2016.

\bibitem{mehta2018espnet}
S.~Mehta, M.~Rastegari, A.~Caspi, L.~Shapiro, and H.~Hajishirzi, ``Espnet:
  Efficient spatial pyramid of dilated convolutions for semantic
  segmentation,'' in \emph{Proceedings of the european conference on computer
  vision (ECCV)}, 2018, pp. 552--568.

\bibitem{mehta2019espnetv2}
S.~Mehta, M.~Rastegari, L.~Shapiro, and H.~Hajishirzi, ``Espnetv2: A
  light-weight, power efficient, and general purpose convolutional neural
  network,'' in \emph{Proceedings of the IEEE/CVF Conference on Computer Vision
  and Pattern Recognition}, 2019, pp. 9190--9200.

\bibitem{wang2019lednet}
Y.~Wang, Q.~Zhou, J.~Liu, J.~Xiong, G.~Gao, X.~Wu, and L.~J. Latecki, ``Lednet:
  A lightweight encoder-decoder network for real-time semantic segmentation,''
  in \emph{2019 IEEE International Conference on Image Processing
  (ICIP)}.\hskip 1em plus 0.5em minus 0.4em\relax IEEE, 2019, pp. 1860--1864.

\bibitem{liu2020fddwnet}
J.~Liu, Q.~Zhou, Y.~Qiang, B.~Kang, X.~Wu, and B.~Zheng, ``Fddwnet: a
  lightweight convolutional neural network for real-time semantic
  segmentation,'' in \emph{ICASSP 2020-2020 IEEE International Conference on
  Acoustics, Speech and Signal Processing (ICASSP)}.\hskip 1em plus 0.5em minus
  0.4em\relax IEEE, 2020, pp. 2373--2377.

\bibitem{zhao2018icnet}
H.~Zhao, X.~Qi, X.~Shen, J.~Shi, and J.~Jia, ``Icnet for real-time semantic
  segmentation on high-resolution images,'' in \emph{Proceedings of the
  European conference on computer vision (ECCV)}, 2018, pp. 405--420.

\bibitem{poudel2019fast}
R.~P. Poudel, S.~Liwicki, and R.~Cipolla, ``Fast-scnn: Fast semantic
  segmentation network,'' \emph{arXiv preprint arXiv:1902.04502}, 2019.

\bibitem{gu2019net}
Z.~Gu, J.~Cheng, H.~Fu, K.~Zhou, H.~Hao, Y.~Zhao, T.~Zhang, S.~Gao, and J.~Liu,
  ``Ce-net: Context encoder network for 2d medical image segmentation,''
  \emph{IEEE transactions on medical imaging}, vol.~38, no.~10, pp. 2281--2292,
  2019.

\bibitem{saini2020ulsam}
R.~Saini, N.~K. Jha, B.~Das, S.~Mittal, and C.~K. Mohan, ``Ulsam:
  Ultra-lightweight subspace attention module for compact convolutional neural
  networks,'' in \emph{Proceedings of the IEEE/CVF Winter Conference on
  Applications of Computer Vision}, 2020, pp. 1627--1636.

\bibitem{xie2018vortex}
C.-W. Xie, H.-Y. Zhou, and J.~Wu, ``Vortex pooling: Improving context
  representation in semantic segmentation,'' \emph{arXiv preprint
  arXiv:1804.06242}, 2018.

\bibitem{arbelaez2010contour}
P.~Arbelaez, M.~Maire, C.~Fowlkes, and J.~Malik, ``Contour detection and
  hierarchical image segmentation,'' \emph{IEEE transactions on pattern
  analysis and machine intelligence}, vol.~33, no.~5, pp. 898--916, 2010.

\bibitem{zheng2015conditional}
S.~Zheng, S.~Jayasumana, B.~Romera-Paredes, V.~Vineet, Z.~Su, D.~Du, C.~Huang,
  and P.~H. Torr, ``Conditional random fields as recurrent neural networks,''
  in \emph{Proceedings of the IEEE international conference on computer
  vision}, 2015, pp. 1529--1537.

\bibitem{yu2016multi}
F.~Yu and V.~Koltun, ``Multi-scale context aggregation by dilated convolutions
  (2015),'' \emph{arXiv preprint arXiv:1511.07122}, 2016.

\bibitem{zhou2020aglnet}
Q.~Zhou, Y.~Wang, Y.~Fan, X.~Wu, S.~Zhang, B.~Kang, and L.~J. Latecki,
  ``Aglnet: Towards real-time semantic segmentation of self-driving images via
  attention-guided lightweight network,'' \emph{Applied Soft Computing},
  vol.~96, p. 106682, 2020.

\bibitem{dong2020real}
G.~Dong, Y.~Yan, C.~Shen, and H.~Wang, ``Real-time high-performance semantic
  image segmentation of urban street scenes,'' \emph{IEEE Transactions on
  Intelligent Transportation Systems}, vol.~22, no.~6, pp. 3258--3274, 2020.

\bibitem{kingma2014adam}
D.~P. Kingma and J.~Ba, ``Adam: A method for stochastic optimization,''
  \emph{arXiv preprint arXiv:1412.6980}, 2014.

\bibitem{pytorch}
\BIBentryALTinterwordspacing
A.~Paszke, S.~Gross, S.~Chintala, G.~Chanan, E.~Yang, Z.~DeVito, Z.~Lin,
  A.~Desmaison, L.~Antiga, and A.~Lerer, ``Automatic differentiation in
  pytorch,'' in \emph{NIPS 2017 Workshop on Autodiff}, 2017. [Online].
  Available: \url{https://openreview.net/forum?id=BJJsrmfCZ}
\BIBentrySTDinterwordspacing

\bibitem{bulo2018place}
S.~R. Bulo, L.~Porzi, and P.~Kontschieder, ``In-place activated batchnorm for
  memory-optimized training of dnns,'' in \emph{Proceedings of the IEEE
  Conference on Computer Vision and Pattern Recognition}, 2018, pp. 5639--5647.

\bibitem{bilinski2018dense}
P.~Bilinski and V.~Prisacariu, ``Dense decoder shortcut connections for
  single-pass semantic segmentation,'' in \emph{Proceedings of the IEEE
  Conference on Computer Vision and Pattern Recognition}, 2018, pp. 6596--6605.

\bibitem{chen2014semantic}
L.-C. Chen, G.~Papandreou, I.~Kokkinos, K.~Murphy, and A.~L. Yuille, ``Semantic
  image segmentation with deep convolutional nets and fully connected crfs,''
  \emph{arXiv preprint arXiv:1412.7062}, 2014.

\bibitem{hu2020temporally}
P.~Hu, F.~Caba, O.~Wang, Z.~Lin, S.~Sclaroff, and F.~Perazzi, ``Temporally
  distributed networks for fast video semantic segmentation,'' in
  \emph{Proceedings of the IEEE/CVF Conference on Computer Vision and Pattern
  Recognition}, 2020, pp. 8818--8827.

\bibitem{zhang2019customizable}
Y.~Zhang, Z.~Qiu, J.~Liu, T.~Yao, D.~Liu, and T.~Mei, ``Customizable
  architecture search for semantic segmentation,'' in \emph{Proceedings of the
  IEEE/CVF Conference on Computer Vision and Pattern Recognition}, 2019, pp.
  11\,641--11\,650.

\bibitem{lin2020graph}
P.~Lin, P.~Sun, G.~Cheng, S.~Xie, X.~Li, and J.~Shi, ``Graph-guided
  architecture search for real-time semantic segmentation,'' in
  \emph{Proceedings of the IEEE/CVF Conference on Computer Vision and Pattern
  Recognition}, 2020, pp. 4203--4212.

\bibitem{9292449cgnet}
T.~Wu, S.~Tang, R.~Zhang, J.~Cao, and Y.~Zhang, ``Cgnet: A light-weight context
  guided network for semantic segmentation,'' \emph{IEEE Transactions on Image
  Processing}, vol.~30, pp. 1169--1179, 2021.

\bibitem{yang2020small}
Z.~Yang, H.~Yu, M.~Feng, W.~Sun, X.~Lin, M.~Sun, Z.-H. Mao, and A.~Mian,
  ``Small object augmentation of urban scenes for real-time semantic
  segmentation,'' \emph{IEEE Transactions on Image Processing}, vol.~29, pp.
  5175--5190, 2020.

\end{thebibliography}
\end{document}